\pdfoutput=1

\documentclass[11pt]{article}

\usepackage[preprint]{acl}

\usepackage{times}
\usepackage{latexsym}
 
\usepackage[T1]{fontenc}

\usepackage[utf8]{inputenc}

\usepackage{microtype}

\usepackage{inconsolata}

\usepackage{graphicx}
\usepackage{multirow}
\usepackage{amsmath}
\usepackage{amssymb}
\usepackage{booktabs}
\usepackage{subcaption}
\usepackage{bm}

\usepackage{cleveref}

\usepackage{color}
\title{Detecting Subtle Differences between Human and Model Languages Using Spectrum of Relative Likelihood}

\author{
 \textbf{Yang Xu\textsuperscript{1,*}},
 \textbf{Yu Wang\textsuperscript{2}},
 \textbf{Hao An\textsuperscript{1}},
 \textbf{Zhichen Liu\textsuperscript{1}},
 \textbf{Yongyuan Li\textsuperscript{1}}
 \\
 \textsuperscript{1}Department of Computer Science and Engineering, \\ Southern University of Science and Technology \\
 \textsuperscript{2}Digital Linguistics Lab, Department of Linguistics,\\ Bielefeld University \\
 \small{
   \textbf{*Correspondence:} \href{mailto:xuyang@sustech.edu.cn}{xuyang@sustech.edu.cn}
 }
}

\begin{document}
\maketitle
\begin{abstract}

Human and model-generated texts can be distinguished by examining the magnitude of likelihood in language. However, it is becoming increasingly difficult as language model's capabilities of generating human-like texts keep evolving. 
This study provides a new perspective by using the relative likelihood values instead of absolute ones, and extracting useful features from the spectrum-view of likelihood for the human-model text detection task. 
We propose a detection procedure with two classification methods, supervised and heuristic-based, respectively, which results in competitive performances with previous zero-shot detection methods and a new state-of-the-art on short-text detection. 
Our method can also reveal subtle differences between human and model languages, which find theoretical roots in psycholinguistics studies. Our code is available at \url{https://github.com/CLCS-SUSTech/FourierGPT}.
\end{abstract}

\section{Introduction}
A recent focus in natural language generation is to develop effective methods of detecting model-generated texts from real human texts. 
This endeavor gives human users an opportunity to know whether a given text, e.g., an explanation, is either generated by language model or written by human, allowing them to decide how much trust to place in the explanation. 
The current most effective methods for this task utilize the likelihood information in text data (e.g.,see \citealp[]{verma-etal-2024-ghostbuster, venkatraman-etal-2024-gpt}). More specifically, most existing methods rely on the observation that the \emph{absolute} likelihood values of texts naturally distribute differently, depending on their sources. 
Examples include the early work like GLTR \citep{gehrmann-etal-2019-gltr} and more recent ones like DetectGPT \citep{mitchell2023detectgpt} and Fast-DetectGPT \citep{bao2024fast}.

However, existing methods have the following limitations: \underline{First}, most work estimates likelihood as a \textbf{static} property of language, but overlooks the fact that human language processing is a \textbf{dynamic} process during which the likelihood of information under processing is bound to certain linguistic and cognitive constraints. For example, the trade-off between processing effort and likelihood of words \citep{levy2008expectation, smith2013effect}, limited and selective context access \citep{gibson1998linguistic, gibson2000dependency}, activation decay \cite{lewis2006computational} and so on. 
\underline{Second}, merely using the \emph{absolute} likelihood values to distinguish generated and human texts is a tricky ``cat and mouse game'' -- as models' capability of mimicking human language keeps growing, their productions would eventually become hardly distinguishable \citep{NEURIPS2020_1457c0d6}. 
\underline{Third}, current methods are not computationally economical, because most of them need to run at least one time of inference on text with a fairly large language model.

We propose a human-model text detection approach that addresses the aforementioned issues and achieves better or comparable performance with existing methods.
Our approach extracts features from the spectrum view of relative likelihood scores of texts, to capture the dynamic changes of likelihood in language. These features are used to design two types of classifiers, a supervised learning-based one and a heuristic-based zero-shot one, both of which reach impressive performances.
The core idea is to obtain the spectrum of likelihood using the Fourier transform, which summarizes the complex patterns of likelihood change in time domain into a much more compact view that magnifies the subtle differences between different texts. 
It has a theoretical basis in psycholinguistic studies on surprisal (likelihood) distribution in natural language. 
Further, our method is still effective when likelihood scores are estimated by na\"ive \textit{n}-gram models, which places much less computational cost.
We name our approach FourierGPT, inspired by existing methods like DetectGPT and Fast-DetectGPT. 


\section{Related Work}
\subsection{Likelihood-based zero-shot text detection}\label{sec:related_work1}
``Zero-shot'' means the text detection is cast not as a supervised classification task, but rather a statistics-based detection task. 
Early works directly use the magnitude of token-level likelihood scores. For example, \citet{gehrmann-etal-2019-gltr} renders the likelihood value of tokens to human-readable color schemes, which creates good visual distinction between GPT-2 generated text and real human texts.
Token ranks based on their log-likelihood scores (LogRank) are used for the same task \citep{solaiman2019release}. 

Recent works develop more advanced statistics based on deeper insights into the distributional difference between human-created and model-generated languages in log-likelihood space. 
For example, \citet{mitchell2023detectgpt} finds that the probability distribution of model-generated text tends to lie under the areas of negative curvature of the log-likelihood function, and in contrast, human text tends not.
Based on this finding, they propose DetectGPT, a zero-shot detection method that measures \emph{perturbation discrepency}, the gap between an original text and its rewritten variant that maintains the same meaning. The assumption is that human text presents smaller gaps than model text.
\citet{bao2024fast} make substantial methodological improvements to DetectGPT and propose Fast-DetectGPT by replacing the probability curvature with conditional probability curvature, which broadly improves the detection accuracy and greatly shorten the computational time.
Therefore, Fast-DetectGPT is the main state-of-the-art method compared with in this study. 
In nature, both DetectGPT and Fast-DetectGPT find an empirical threshold for the variance of \emph{absolute} likelihood values, which depend on the choice of the inference model.

Two other likelihood-based methods are also compared within this study: normalized log-rank perturbation (NPR) \citep{su2023detectllm} and divergence between multiple completions of a truncated passage (DNA-GPT) \citep{yang2023dna}. 
Both rely on estimation of absolute likelihood to some extent. 

\subsection{Surprisal and likelihood of language}
The way likelihood scores are defined in the previous section is equivalent to the concepts of ``surprisal'' and ``information density'', which are commonly used interchangeably in the psycholinguistics literature.
Surprisal is known to reflect the cognitive load of processing a word, phrase, or sentence -- it takes more effort and time to produce and comprehend units of higher surprisal, such as rare words \cite{hale2001probabilistic}.
There is a preference in human language to keep the surprisal intensity evenly distributed in time, known as uniform information density (UID) \citep{jaeger2010redundancy}, or entropy rate constancy (ERC) \citep{genzel2002entropy, genzel2003variation}.
This preference is an outcome of the speaker/writer's intention to make the listener's comprehension easier, which, therefore, draws a potential connection to the topic of this study -- is this preference learned by language models?
Another relevant work is \citet{xu2017spectral}'s finding that periodicity of surprisal exists in natural language, which can be captured by spectrum analysis methods and be used to predict the interaction outcome of dialogue partners. 

Understanding the human mind's preference and tendency in handling surprisal/likelihood leads to new ideas for natural language generation techniques.
For example, some recent endeavors build on top of the assumption that model-generated language appears more natural and human-like if it is generated through a decoding algorithm that follows the UID theory, such as the beam search algorithm as evidenced in \citet{meister2020if}; or it falls under the so-called stable entropy zone \citep{arora2023stable};
\citet{meister2023locally} proposes locally typical sampling, which enforces the uniform distribution of likelihood during the generation process, and results in generated texts that are more aligned with human texts.


\subsection{Evaluation of natural language generation with likelihood}

The task of evaluating natural language generation (NLG) is essentially related to the text detection task. 
Therefore, likelihood (and its variants) is a natural option here. 
Early works in NLG often frame the evaluation equivalent to a detection task, which treats human text as gold-standard, and uses the ``distance'' from human text to measure the quality of generated text.
For example, \citet{ippolito-etal-2020-automatic} uses total probability as a measurement 
and \citet{Holtzman2020The} compares the generation perplexity and Zipf coefficient \citep{zipf1949human} (closely related to LogRank) of texts from different sampling methods. 
These works are very similar in methodology to those reviewed in \Cref{sec:related_work1}, only except that they did not emphasize detection accuracy, but focused on ``quality control'' of generation. 

Some recent evaluation metrics compare model text with human text in high dimensional space, such as MAUVE \citep{pillutla2021mauve}. While this type of method does not directly use likelihood information, interesting correlations have been found between likelihood-based metrics.
For example, \citet{yang2023face} proposes a novel evaluation metric called Fourier analysis of cross-entropy (FACE), which converts the cross-entropy scores (i.e., likelihood) to spectrum representations and then measures the distances in frequency-domain. The resulted distance scores can reflect generation qualities that are co-examined with other metrics, such as MAUVE, and align well with human judgements. This work indicates that with proper transformation on simple likelihood scores, rich insights about language use are viable. 
In fact, the text detection method proposed in this study is directly inspired by \citeauthor{yang2023face}'s work \citeyearpar{yang2023face}. 

\section{Method}
\begin{figure*}[h]
\centering
\includegraphics[width=.95\linewidth]{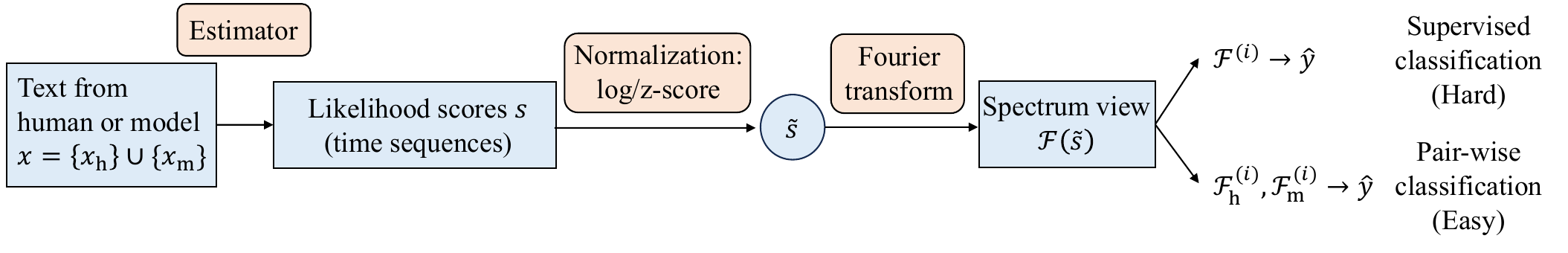}
\includegraphics[width=.95\linewidth]{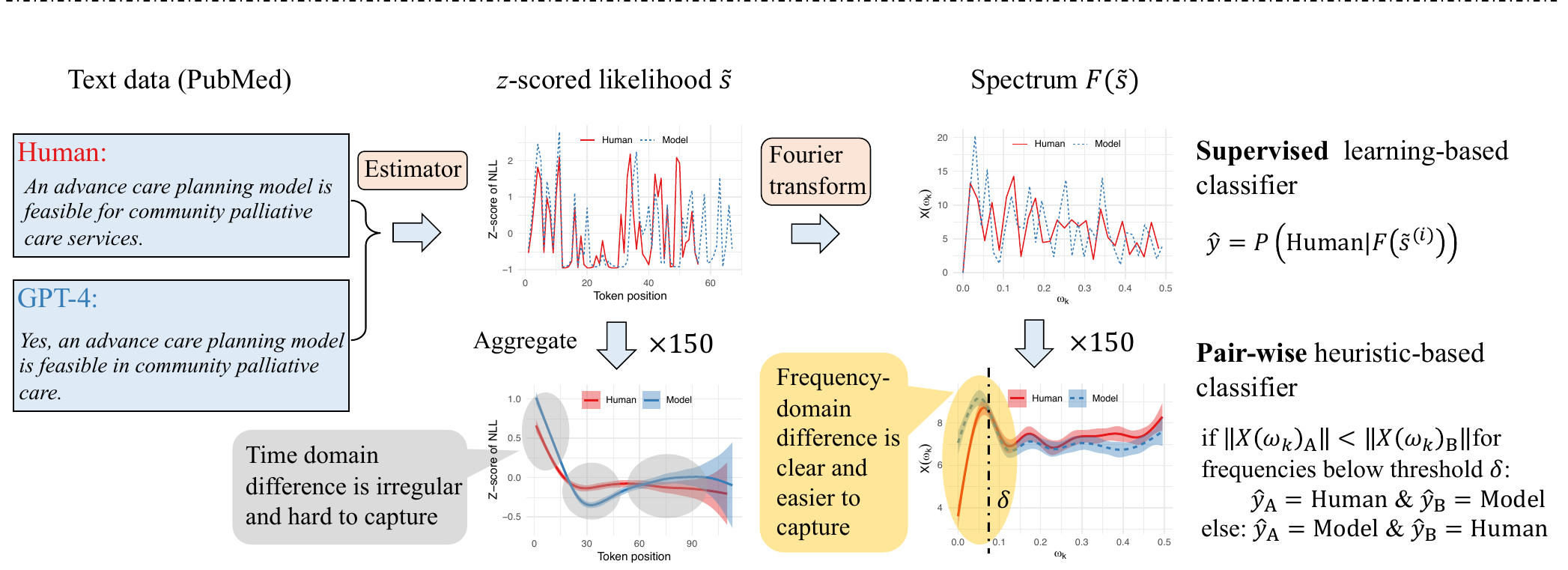}
\caption{The procedure (above) and example (below) of FourierGPT.}
\label{fig:procedure}
\end{figure*}

The procedure of FourierGPT consists of three steps: 1) Estimate and normalize likelihood scores; 2) Carry out Fourier transform to get the spectrum view; 3) Conduct classification on the spectrum.
The procedure is illustrated in \Cref{fig:procedure} with an example. Details of each step are described below.

\subsection{Estimation and normalization of likelihood scores}
We estimate the likelihood scores of text data with pretrained language models of various scales: Mistral-7B \citep{jiang2023mistral}, GPT-2 families \citep{radford2019language}, and a bigram language model trained from scratch. 
We use the implementation of bigram model from \citet{kenlm} and train it on a subset of C4 dataset \citep{JMLR:v21:20-074}.
Raw likelihood scores are estimated by running a forward pass on the input text. Taking an input of $N$ tokens, $t_1,\dots,t_N$, the likelihood score of $i$th token $s_i$ is computed by $s_i = -\log P(t_i|t_1,\dots,t_{i-1})$, i.e., the negative log probability returned by the language model, which we call the \emph{estimator} model.


Then we normalize the raw likelihood scores $s_1,\dots,s_N$ within each sequence, obtaining the $z$-scored likelihood $\tilde{s}_1,\dots,\tilde{s}_N$, in which $\tilde{s}_i = \frac{s_i - \mu}{\sigma}$, $\mu = \sum s_i/N$ and $\sigma = \sum (s_i-\mu)^2/(N-1)$. 
Here we would like to stress that this seemingly trivial normalization step is actually \emph{critical} for verifying the hypothesis of this study: the raw likelihood $s_i$'s value depends on the choice of estimator -- larger models usually result in smaller values (similar to perplexity); but the $z$-scored $\tilde{s}_i$ characterizes the relative level of likelihood within the range $(-\infty,\infty)$, which is less dependent on the estimator. 
As expected, the distribution of raw likelihood is highly skewed, while the $z$-scored likelihood is closer to normal distribution and gives better classification results in the following steps.  


We also find that $z$-score normalization before the Fourier transform is supported by practices in signal processing. \citet{reno2018quantification} points out the secondary motion imaging artifact: when a time series consists of high and low-frequency components of different intensities, the vanilla Fourier transformed spectrum will be dominated by the low-frequency component, and normalization can eliminate this artifact.

\subsection{Fourier transform}
The next step is to obtain the spectrum view of the $z$-scored likelihood sequence $\tilde{s}_0,\dots,\tilde{s}_{N-1}$ as input. We apply discrete Fourier transform (DFT) according to the following:

\begin{equation}\label{eq:DFT}
    X(\omega_k)\triangleq \sum_{n=0}^{N-1}\tilde{s}_n e^{-j\omega_k n} 
\end{equation}

The result is a set of complex numbers $\mathcal{F}=\{X(\omega_k)\}|_{k=0,\dots, N-1}$ as the frequency-domain representation of the input time-domain signal (likelihood scores), in which $\omega_k$ is the $k$-th frequency component. 
We change the starting index of $\tilde{s}_i$ to 0 because DFT requires $k=0$ as the lowest frequency component. 
$X(\omega_k)$ is a complex number made up of real and imaginary parts, $X(\omega_k)=\text{Re}(X(\omega_k))+\text{Im}(X(\omega_k))j$. The norm $\lVert X(\omega_k)\rVert=\sqrt{\text{Re}(X(\omega_k))^2 + \text{Im}(X(\omega_k))^2}$ represents the intensity of the $k$th component $\omega_k$.
Finally, we use the sequence $\{\lVert X(\omega_k)\rVert\}|_{k=0,\dots, N-1}$ as the \textbf{spectrum-view of likelihood}, which provides features for the next classification step. 

The range of $\omega_k$ is $[0,\pi]$, and its interpretation is not trivial. Based on an intuitive interpretation provided by \citet{yang2023face}, we can roughly tell that the likelihood score $\tilde{s}_i$ at the level of $\lVert X(\omega_k)\rVert$ tends to occur every $1/\omega_k$ tokens in the text data. 
Interestingly, we find the way $\lVert X(\omega_k)\rVert$ distributes along $\omega_k$ provides unique information to distinguish human from model. To develop solid explanations of what the spectrum of likelihood means is important, yet a different topic. We primarily focus on \emph{harnessing} the spectrum information for the detection task, and try to do gain some interpretive insights at our best in \Cref{sec:discussion}. 

\subsection{Classification methods}
We use two classification methods for the text detection task: A \textbf{supervised} learning-based classifier trained from the entire labeled spectrum data, which makes a binary prediction (human or model) on any given input spectrum representation; and a \textbf{pair-wise} heuristic-based classifier that tells which one is from human (hence the other one is from a model) in any given \emph{pair} of input spectrum representations. For the pair-wise classifier, we require that the input pair must come from the same text \emph{prompt}, which guarantees that one of them is from human and the other one is from model. 
It is obvious to see that the supervised classifier is more difficult to train as no prior information is given. 

\vspace{-.1em}
\subsubsection{Supervised learning-based classifier}
We train the supervised classifier using an augmented spectrum as input feature, which is obtained with multiple rounds of  \textit{circularization} operation on the likelihood scores: 
given an original time series of likelihood scores $\mathcal{C}_0=s_1,s_2,\dots,s_n$, circularization at step $T$ is to chop off the segment of length $T$ at the head and then append it to the end, resulting in a new series $\mathcal{C}_T=s_{T+1},\dots,s_n,s_1,\dots,s_{T}$. See the following complete procedure:

\vspace{-1em}
\begin{equation*}
\boxed{
\begin{array}{rc}
\text{Original scores } \mathcal{C}_0 &\rightarrow s_1,s_2,\dots,s_n\\
\text{Circularized scores } \mathcal{C}_1 &\rightarrow s_2,\dots,s_n,s_1\\
\text{Circularized scores } \mathcal{C}_2 &\rightarrow s_3,\dots,s_1,s_2\\
\vdots & \\
\text{Circularized scores } \mathcal{C}_{n-1} &\rightarrow s_{n},s_1,\dots,s_{n-1}
\end{array}
}
\end{equation*}

Next, we apply Fourier transform to each circularized likelihood sequence, which produces $n$ spectra in total, $\mathcal{F}(\mathcal{C}_t)$, $t=0,\dots,n-1$. 
The average spectrum $\overline{\mathcal{F}}=\frac{1}{n}\sum \mathcal{F}(\mathcal{C}_t)$ is used as the input feature for training the classifier. 
Lastly, we train several common types of classification models and evaluate their performances in \Cref{sec:supervised_cls_res}. 
The circularization operation draws inspiration from the circular convolution in digital signal processing \citep{elliott2013handbook}.
The intuition is: if a weak periodicity exists in the original ``signal'' $\mathcal{C}_0$, then obtaining multiple spectra from its multiple variants ($\mathcal{C}_1$ through $\mathcal{C}_{n-1}$) should amplify the periodicity that is undetectable otherwise.
From a machine learning perspective, it is like a way of data augmentation, which picks the most salient features by aggregating multiple variants of the original data. 


\subsubsection{Pair-wise Heuristic-based classifier}
\begin{figure}[ht]
\centering
\begin{subfigure}{\linewidth}
	\centering
	\includegraphics[width=\linewidth]{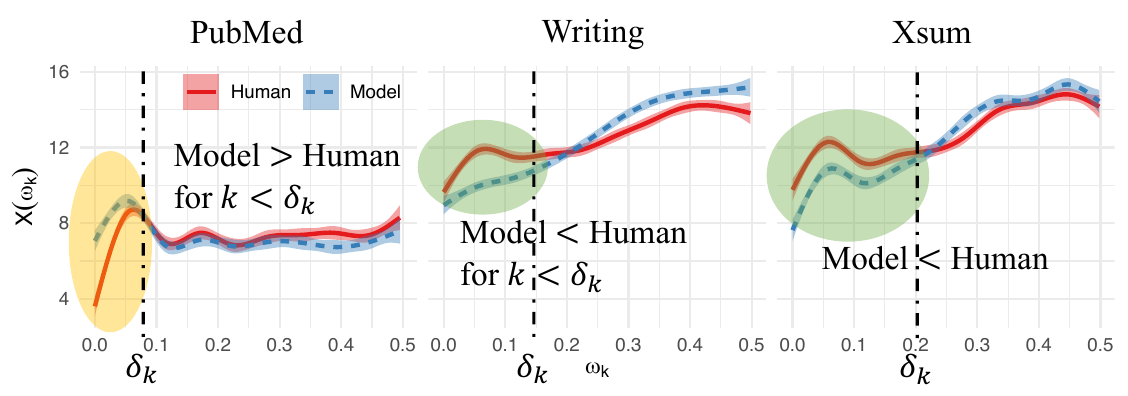}
	\caption{Human and GPT-4}
\end{subfigure}
\begin{subfigure}{\linewidth}
	\centering
	\includegraphics[width=\linewidth]{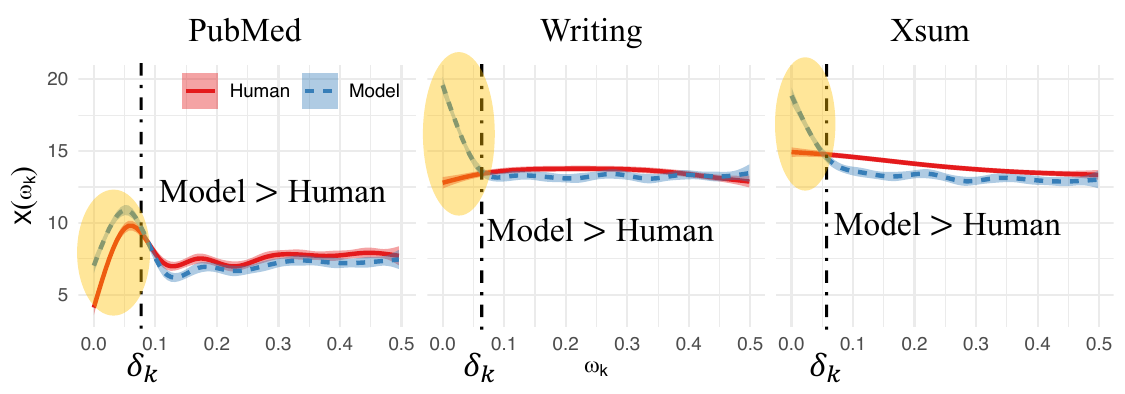}
	\caption{Human and GPT-3.5}
\end{subfigure}
\begin{subfigure}{\linewidth}
	\centering
	\includegraphics[width=\linewidth]{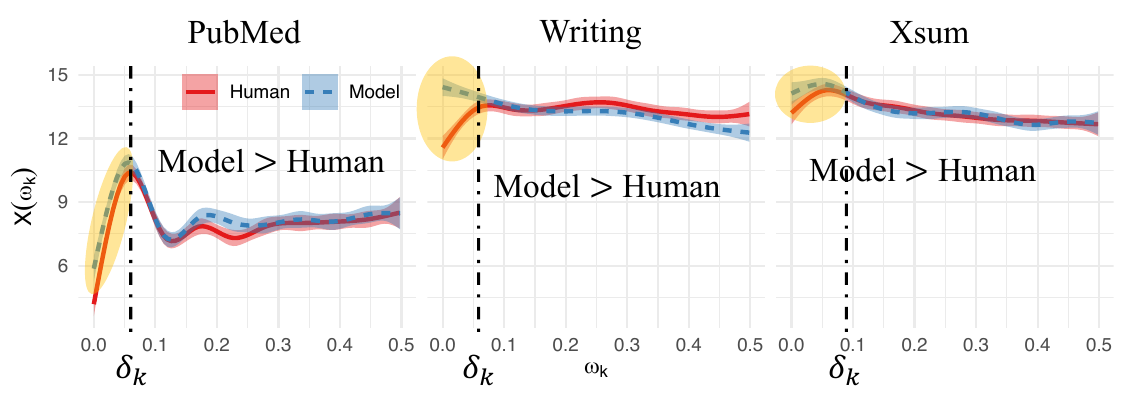}
	\caption{Human and GPT-3}
\end{subfigure}
\caption{Heuristics for constructing pair-wise classifiers: Likelihood spectrum shows salient difference at low frequency components. Curves are fit using generative additive models (GAM). Shaded areas are 95\% confidence intervals from bootstrap.}
\label{fig:heuristics}
\end{figure}

We design a set of classifiers based on an empirical \emph{heuristic} obtained by observing the difference between human and model's spectrum views: the likelihood spectrum presents a salient \textbf{difference at the low-frequency end}. 
The direction of difference slightly varies across dataset $\times$ model groups, but for most groups, the model's spectrum has a larger power amplitude than the human's, except for GPT-4 on Writing and Xsum (see \Cref{fig:heuristics}).

The heuristic is expressed as follows:

\vspace{-1em}
\begin{equation*}
	\left| \sum_{k=1}^{\delta_k}\lVert X^{\text{Human}}(\omega_k) \rVert - \sum_{k=1}^{\delta_k}\lVert X^{\text{Model}}(\omega_k) \rVert \right| > \varepsilon
\end{equation*}

in which $\delta_k \in \mathbb{Z}$ is an integer threshold defining the range of frequency components $\omega_k$ selected for comparing the spectrum power $\lVert X(\omega_k) \rVert$ s.t. $1\leq k\leq \delta_k$, and its value is determined empirically in each dataset group. 
$\varepsilon\in \mathbb{R}$ is a real number threshold characterizing the observed difference in $\lVert X(\omega_k) \rVert$ between human and model, which is also determined empirically. A larger $\varepsilon$ value means a more strict standard for distinguishing $\lVert X^{\text{Human}}(\omega_k) \rVert$ and $\lVert X^{\text{Model}}(\omega_k) \rVert$. 
In our experiments, we use $\varepsilon=0$ for convenience.

\section{Experiment Results}

\subsection{Datasets}
We use the text detection datasets provided by \citet{bao2024fast}, which follows the experiment settings of \citet{mitchell2023detectgpt}. It makes sure that all comparisons made to the previous methods are valid and consistent.
The datasets text prompts are gathered from three sources: PubMedQA dataset \citep{jin2019pubmedqa} which consists of 273.5k human experts' answers to biomedical research questions; Reddit WritingPrompts dataset \citep{fan-etal-2018-hierarchical} which includes 300k human-written stories with prompts; XSum dataset \cite{narayan-etal-2018-dont} which contains human-written summarization of 226.7k online articles in British Broadcasting Corporation (BBC).
The datasets are compiled by \citet{bao2024fast}, using the OpenAI API\footnote{\url{https://openai.com/blog/openai-api}}. Three APIs are used for generation: GPT-4, GPT-3.5 (ChatGPT), and GPT-3 (Davinci).
Each one of the three datasets (PubMed, Writing, and Xsum) contains 150 pairs of human and model texts. Each pair shares the first 30 tokens and differs afterward.
Therefore, our main experiments work on 3 (genres) $\times$ 3 (generation models) $=9$ conditions.

\begin{table}[t]
\centering
\begin{tabular}{l l c c}
\toprule
\textbf{Dataset} & \textbf{Gen. model} & \textbf{Best acc.} & \textbf{Classifier} \\
\midrule
\multirow{3}*{PubMed} & GPT-4 & \textbf{0.8267} & SVM \\
& GPT-3.5 &  0.6000 & SVM\\
& GPT-3 & 0.5800 & HGBT \\
\midrule
\multirow{3}*{Writing} & GPT-4 & 0.7167 & NB \\
& GPT-3.5 & 0.7500 & NB \\
& GPT-3 & \textbf{0.8267} & SVM \\
\midrule
\multirow{3}*{Xsum} & GPT-4 & 0.7400 & SVM \\
& GPT-3.5 & 0.7500 & SVM \\
& GPT-3 & 0.7400 & SVM \\
\bottomrule
\end{tabular}
\caption{Accuracy scores of supervised classifiers. All spectrum features used for training are based on likelihood scores estimated by GPT2-xl model. }
\label{tab:supervised_cls}
\end{table}

\begin{table}[ht]
\centering
\begin{tabular}{l c c c}
\toprule
\multirow{2}*{\textbf{Method}} & \textbf{PubMed} & \textbf{Writing} & \multirow{2}*{\textbf{Avg.}}\\
& \textbf{GPT-4} & \textbf{GPT-3} & \\
\midrule
Likelihood & 0.8104 & 0.8496 & 0.8300 \\
LogRank & 0.8003 & 0.8320 & 0.8162  \\
DNA-GPT & 0.7565 & 0.8354 & 0.7960 \\
NPR & 0.6328 & 0.7847 & 0.7088 \\
DetectGPT & 0.6805 & 0.7818 & 0.7312 \\
Fast-Detect & \textbf{0.8503} & \textbf{0.9568} & 0.9036 \\
\midrule
FourierGPT & 0.8267$\dagger$ & 0.8267 & 0.8267 \\ 
\bottomrule
\end{tabular}
\caption{Accuracy of our best supervised classifiers compared to other likelihood-based zero-shot methods reported in \citep{bao2024fast} on selected two task subsets. Best scores are in bold, and $\dagger$ indicates second best.}
\label{tab:supervised_cls_pubmed_writing}
\end{table}

\subsection{Supervised learning-based classification}\label{sec:supervised_cls_res}
Six common classification models are trained and evaluated using 5-fold cross-validation, and we find that the Support Vector Machine (SVM) model achieves the best overall performance. The accuracy scores of SVM on all datasets are shown in \Cref{tab:supervised_cls}.
It can be seen that our method performs particularly well for the PubMed dataset, which achieves an above 80\% accuracy score, as compared to the scores around 70\% for the other two datasets. 
This indicates that the supervised classifier can learn features in short texts better than in longer ones.

Although our best scores on PubMed are lower than the state-of-the-art from Fast-DetectGPT (see \Cref{tab:pairwise_cls}), we think this is still an impressive result because it outperforms most of the other previous methods that use \emph{absolute} likelihood scores for detection, and the gap from SOTA is small. 
We list the comparison on PubMed (GPT-4) and Writing (GPT-3) in \Cref{tab:supervised_cls_pubmed_writing}.

\begin{table*}[h!]
\centering
\begin{tabular}{l l c c c | c c c}
\toprule
\textbf{Dataset} & \textbf{Gen. model} & \textbf{FourierGPT} & $\bm{\delta_k}$ & \textbf{Est. model} & \textbf{Fast-Detect} & \textbf{Likelihood} & \textbf{GPTZero} \\
\hline
\multirow{3}*{PubMed} & GPT-4 & \textbf{0.9133} & 3 & GPT2-xl & 0.8503 & 0.8104 & 0.8482 \\
 & GPT-3.5 & \textbf{0.9467} & 2 & Mistral & 0.9021 & 0.8775 & 0.8799 \\
 & GPT-3 & 0.6867 & 5 & Mistral & \textbf{0.7225} & 0.5668 & 0.4246 \\
\hline
\multirow{3}*{Writing} & GPT-4 & 0.8467 & 23 & GPT2-xl & \textbf{0.9612} & 0.8553 & 0.8262 \\
 & GPT-3.5 & 0.9200 & 30 & Mistral & \textbf{0.9916} & 0.9740 & 0.9292 \\
 & GPT-3 & 0.7200 & 6 & Mistral & \textbf{{0.9568}} & 0.8496 & 0.6009 \\
\hline
\multirow{3}*{Xsum} & GPT-4 & 0.8733 & 29 & GPT2-xl & 0.9067 & 0.7980 & \textbf{0.9815} \\
 & GPT-3.5 & 0.9200 & 24 & GPT2-xl & 0.9907 & 0.9578 & \textbf{0.9952} \\
 & GPT-3 & 0.6067 & 13 & GPT2-xl & \textbf{0.9396} & 0.8370 & 0.4860 \\
\bottomrule
\end{tabular}
\caption{Accuracy of pair-wise heuristic-based classifiers. The best accuracy, corresponding heuristic $\delta_k$, and estimator model used are reported. We report the classification accuracy scores from three previous zero-shot text detection methods, including two open-source solutions, Fast-DetecGPT and Likelihood, and one commercial detector GPTZero. We report the scores directly from \citep{bao2024fast}. Best scores are in bold. } 
\label{tab:pairwise_cls}
\end{table*}

\subsection{Pair-wise heuristic-based classification}\label{sec:pairwise_cls_res}
The pair-wise heuristic-based classification results are shown in \Cref{tab:pairwise_cls}.
For a comprehensive comparison, we include the second-best performing open-source method Likelihood and a commercial detection solution GPTZero \citep{tiangptzero} in the table. 
Our method performs generally better on GPT-4 and GPT-3.5 groups: it outperforms the state-of-the-art Fast-DetectGPT on PubMed data, though not as good in Writing or Xsum. Yet, the performance on the latter two datasets is quite competitive to the second-best previous method. 

Similar to the supervised classifier, we also experiment with heuristic-based classifiers using likelihood scores estimated from bigram models, whose accuracy results are shown in \Cref{tab:pairwise_bigram}. It has surprisingly good performance on Writing data: the accuracy on Writing+GPT-3.5 reaches 0.9067, which is better than Fast-DetectGPT.

\begin{table}[ht]
\centering
\begin{tabular}{l l c c}
\toprule
\textbf{Dataset} & \textbf{Best group} & \textbf{Best acc.} & \textbf{Avg. acc.} \\ 
\midrule
Pubmed & GPT-3 & 0.6733 & 0.6511 \\
Writing & GPT-3.5 & \textbf{0.9067} & 0.7867 \\
Xsum & GPT-3.5 & 0.7800 & 0.7289 \\
\bottomrule
\end{tabular}
\caption{Accuracy of FourierGPT pair-wise classifiers using likelihood spectrum from bigram language model. The bold number performs better than Fast-DetectGPT.}
\label{tab:pairwise_bigram}
\end{table}


\section{Discussion: Text Features Affects Spectrum of Likelihood}\label{sec:discussion}
The purpose of this section is to investigate why the spectrum view of relative likelihood scores can be used to distinguish texts from humans and models. What specific features in the text are reflected in the frequency-domain? Can we know more about what language models learned (and did not learn) from humans by reading their likelihood spectrum?
With these questions in mind, we present some interesting patterns discovered.

\begin{table}[h]
\centering
\begin{tabular}{l c c}
\toprule
\textbf{Group} & \textbf{Start w/ ``\textit{Yes}''} & \textbf{Start w/ ``\textit{No}''}\\
\hline
GPT-4 & 78/150 & 10/150\\
GPT-3.5 & 35/150 & 2/150\\
Davinci & 32/150 & 32/150\\
\hline
Human & 0/150 & 0/150\\
\bottomrule
\end{tabular}
\caption{Proportions of answers that start with ``\textit{Yes}''/``\textit{No}'' pattern in PubMed data.}
\label{tab:yesno}
\end{table}

\subsection{Answers starting with ``yes/no''}\label{sec:yesno}
We find that in PubMed data, model-generated answers are much more likely to start with a fixed pattern of ``\textit{Yes}''/``\textit{No}'', while humans do not answer in this style at all (at least in the current data). The ratios of answers with this pattern are listed in \Cref{tab:yesno}. Since each model group comes with a different set of 150 human question/answer texts, the total odds of the human group is as low as 0/450.

This is an interesting finding because it indicates the tendency of models to generate texts of high certainty: when the prompt is in an explicit form like ``\textit{Question: ...}'', then the model tends to address it first by giving a certain answer like ``\textit{Yes}'' or ``\textit{No}''. On the other side, human answers sound less confident and tend to avoid certainty. 
We conjecture that this finding could be due to the general tendency of human language to use more \emph{hedging} and avoid over-confidence, particularly in face of difficult questions such as the highly professional ones in PubMed.

\begin{figure}[h]
\centering
\begin{subfigure}{\linewidth}
    \includegraphics[width=\linewidth]{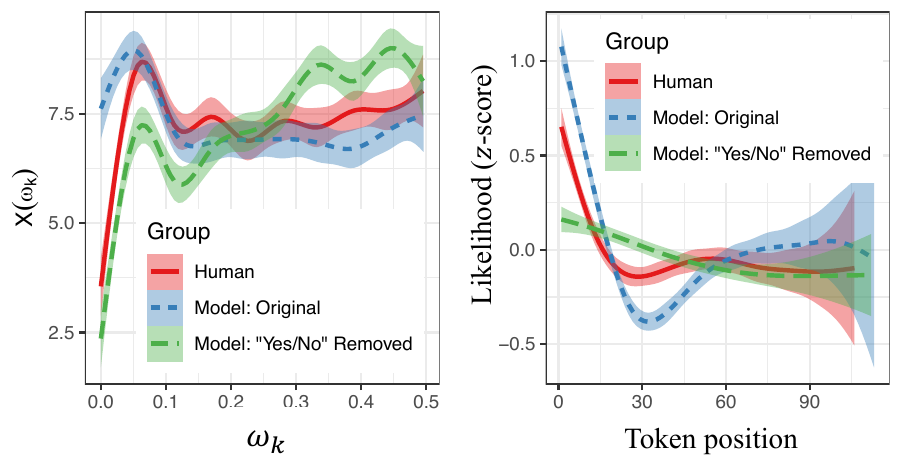}
    \caption{Estimated by GPT-4.}\label{fig:yesno_gpt4}
\end{subfigure}
\begin{subfigure}{\linewidth}
    \includegraphics[width=\linewidth]{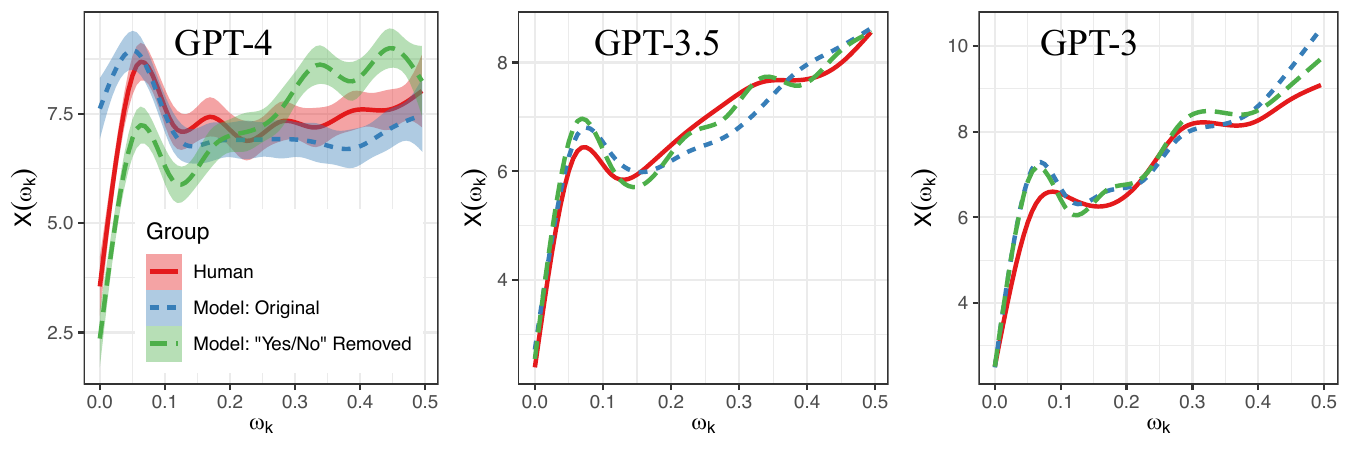}
    \caption{Estimated by all three models.}\label{fig:yesno_all}
\end{subfigure}
\caption{(a) The changes of likelihood spectrum ((a) left) and likelihood-position plot ((a) right) after removing the ``\textit{Yes}''/``\textit{No}'' in answer from PubMed data (estimated by GPT-4 only). (b) The changes of spectrum estimated by all three models.  Curves are fit with GAM. Shaded areas are 95\% confidence intervals from bootstrap. }\label{fig:yesno}
\end{figure}

We use a simple ablation experiment on data to examine whether this subtle difference is reflected in the spectrum of likelihood. We remove the ``\textit{Yes}''/``\textit{No}'' at the beginning of the answer, re-computing the likelihood scores, and re-do the Fourier transform. Consequently, the spectrum of the model morphs in shape towards the direction of human (\Cref{fig:yesno_gpt4} (left)): the altered GPT-4 data's low-frequency components drop, and the high ends rise, both towards the direction of human.
GPT-3.5 and GPT-3 have much fewer data points containing ``\textit{Yes}''/``\textit{No}'', but a similar trend exists in the high end (see \Cref{fig:yesno_all}). 
To showcase the advantage of spectrum view, we also plot the $z$-scored likelihood against token position (\Cref{fig:yesno_gpt4} (right)), which shows that removing ``\textit{Yes}''/``\textit{No}'' makes the likelihood curve flatter (thus, closer to human), but this change is not as easy to describe as the spectrum.
In sum, the subtle differences between human and model languages, like the ``\textit{Yes}''/``\textit{No}'' use discussed here, can be reflected in likelihood space, and the spectrum view can capture this difference conveniently.

\subsection{Text lengths effect}
It is pointed out in previous work that zero-shot detectors are supposed to perform worse on short text because shorter text means fewer data points to compute the likelihood-based statistics \citep{bao2024fast}.
We examine the effect of text length on FourierGPT's performance, by using only the first $n=50,100,150$ tokens for the entire classification procedure on Writing and Xsum datasets. As PubMed data are already short, with the mean length of the answer part being $n=35.2$ words, they are not included in the experiment.

\begin{figure}
\centering
\begin{subfigure}{\linewidth}
	\includegraphics[width=\linewidth]{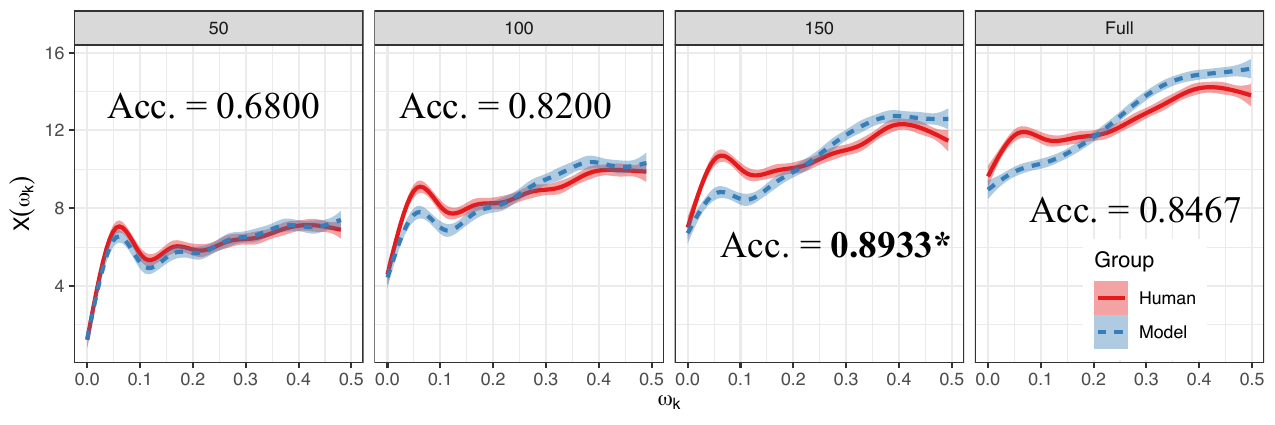}
	\caption{Writing data.}
\end{subfigure}
\begin{subfigure}{\linewidth}
	\includegraphics[width=\linewidth]{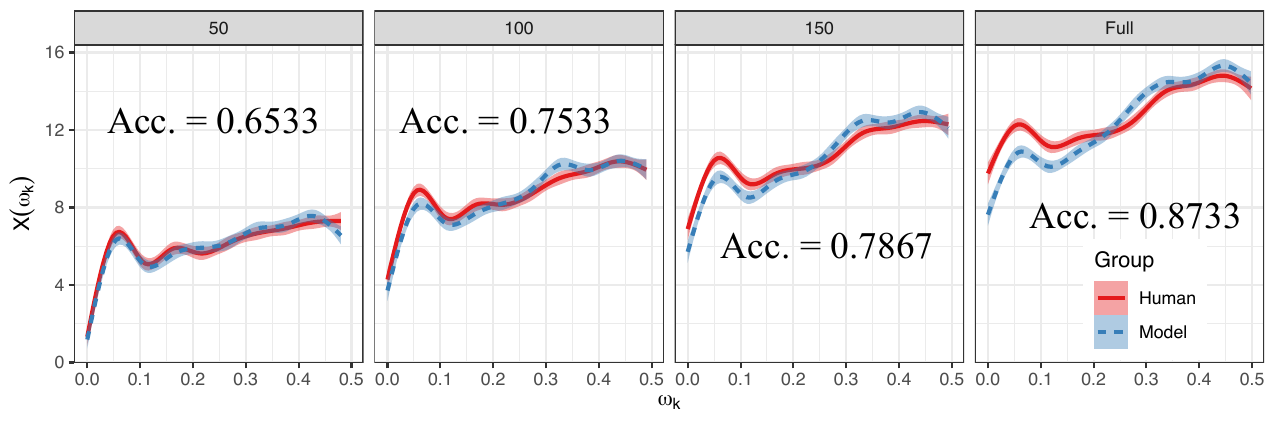}
	\caption{Xsum data.}
\end{subfigure}
\caption{Text lengths affect likelihood spectrum and pairwise classifier performance. Each plot corresponds to lengths of text, $n=50,100,150$, compared to ``Full''. Curves are fit with GAM. Shaded areas are 95\% confidence intervals from bootstrap.}
\label{fig:lengths}
\end{figure}

From \Cref{fig:lengths} it can be seen that shorter texts indeed result in more indistinguishable spectrum shapes between human and model. Surprisingly, however, when we use the cut-off token count $n=150$ on Writing data, the pair-wise classifier's accuracy increases by a significant percentage, even better than using full length.
It strengthens the finding on PubMed that likelihood spectrum better captures the characteristics of short texts.

\subsection{Part-of-speech masking}
As the last part of discussion, we investigate the role played by words of different part-of-speech (POS) tags in affecting the likelihood spectrum. Many previous studies have revealed that adding POS information in either the encoder or decoder side of (multimodal) language models can enhance language understanding and generation tasks (e.g., see \citealp[]{Wray_2019_ICCV,deshpande2019fast}).
\citet{yang-etal-2022-diversifying} further demonstrated that by informing the sampling method of different POS patterns, the generated texts show more diversity without sacrificing quality. 
The above findings suggest that, POS patterns as linguistic knowledge can significantly impact the NLG process.

\begin{figure}[h]
\begin{subfigure}{0.49\linewidth}
    \includegraphics[width=\textwidth]{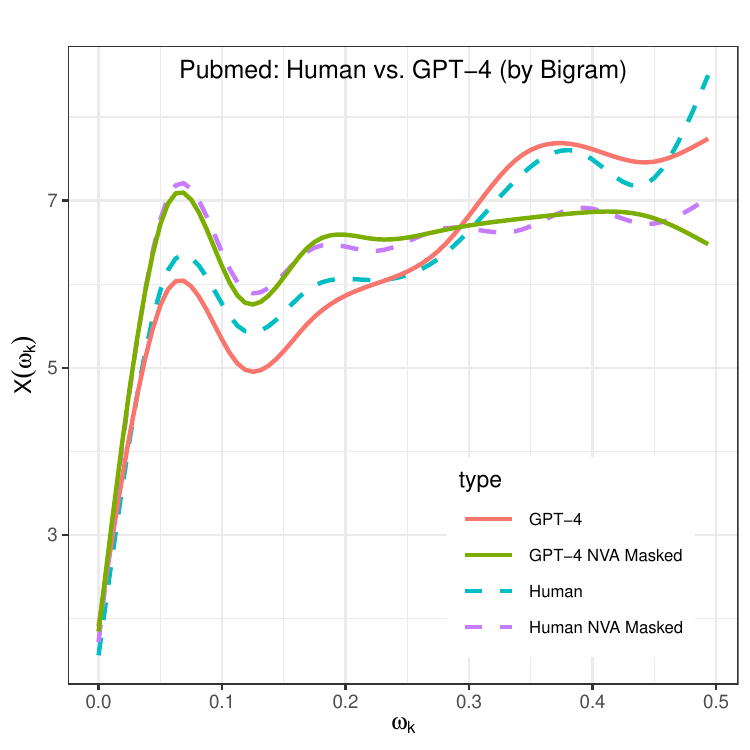}
    \caption{PubMed}
\end{subfigure}
\begin{subfigure}{0.49\linewidth}
    \includegraphics[width=\textwidth]{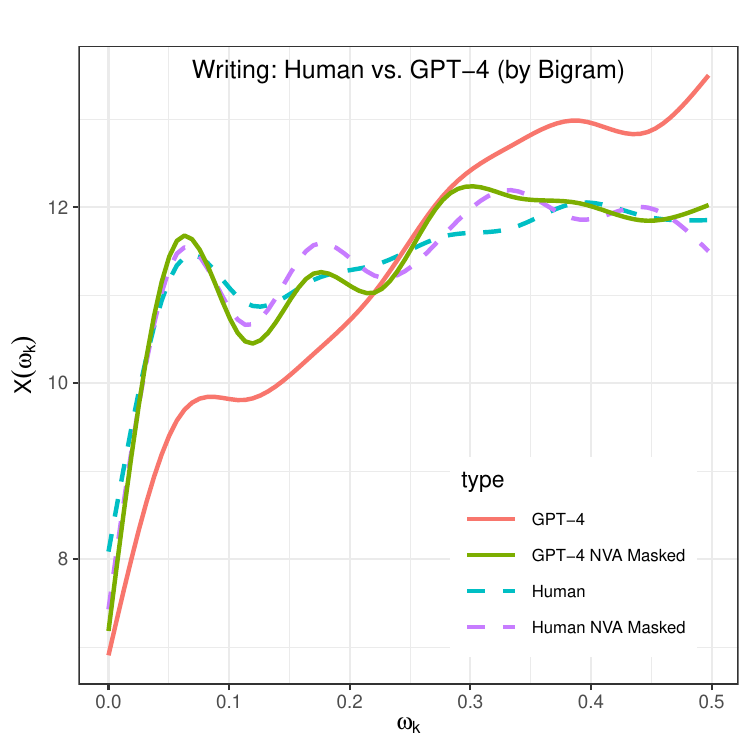}
    \caption{Writing}
\end{subfigure}
\centering
\begin{subfigure}{0.49\linewidth}
    \includegraphics[width=\textwidth]{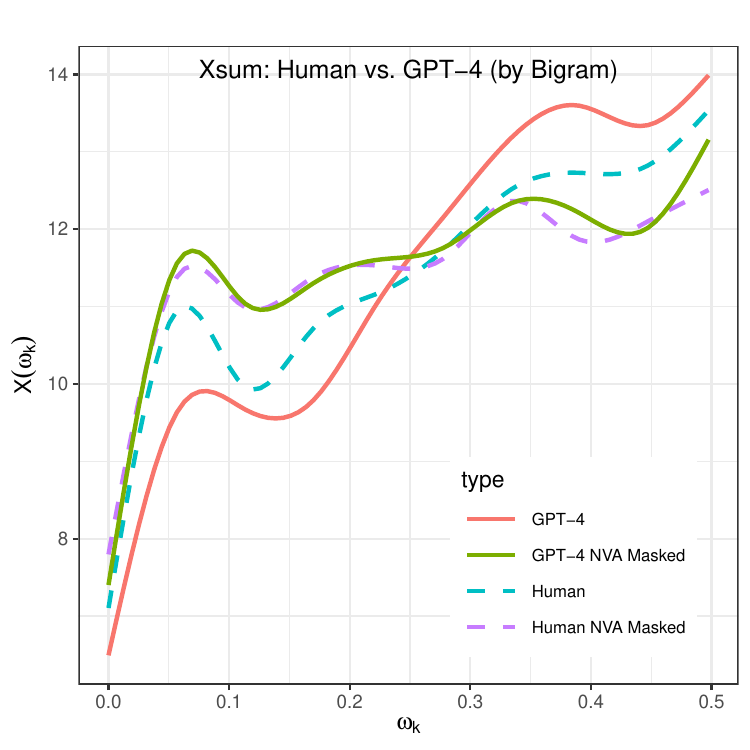}
    \caption{Xsum}
\end{subfigure}
\caption{Likelihood spectrum before and after masking out the `NOUN+VERB+ADJ' POS tags. It can be seen that GPT-4 texts show greater variation before and after the mask, while human text show smaller change. Likelihood is estimated with a bigram model. Curves are fit with GAM. Shaded areas are 95\% confidence intervals from bootstrap.} 
\label{fig:MASK}
\end{figure}

Based on this, we use masking to break some selected POS patterns during the generation process, which will restrain the estimated likelihood from encoding complete information of POS, and then see how this will affect the likelihood spectrum.
First, we mask three POS tags in text: `NOUN', `VERB', and `ADJ', individually; and the union of the three, `NOUN+VERB+ADJ' (thus, `NVA'). 
Then, the masked tokens' likelihood scores are replaced with the average score, thus eliminating the contribution from that specific POS tag. 
Masking is done use the spaCy \citep{Honnibal_spaCy_Industrial-strength_Natural_2020} POS tagger.
The likelihood spectrum results after `NVA' being masked is shown in \Cref{fig:MASK}.


We find an interesting phenomenon: after applying the POS mask, the change of likelihood spectrum for human text is relatively small, while for the model text (GPT-4), the change is much bigger (see \Cref{fig:MASK}).
Such difference is not limited to the collection of `NVA' tags alone, but also observed for the `NOUN', `VERB', and `ADJ' POS tags individually (shown from \Cref{fig:MASK-adj-gpt2} to \Cref{fig:MASK-NVA-gpt2} in Appendix).
This difference is most salient if bigram is used as the estimator\footnote{Technically, it is impossible to apply attention mask in bigram model. Our solution is: as an example, if a verb is encountered, we replace the current likelihood value with a random value between the minimum and maximum, both of which are computed in the same sentence where the verb is located. This operation metaphorically mimics the attention masking.} as compared to GPT-2.
We back up the observation by calculating the \emph{spectral overlap} between the original spectrum and the one after POS masking, for both human and model texts. It is a common metric for measuring spectrum similarity, and it was first used by \citet{yang2023face} as a means for NLG evaluation. 
It turns out that indeed the generated text has smaller before-after spectral overlap, as compared to human text, which indicates that the model spectrum is less stable against POS masking (Details reported in \Cref{tab:speclap} in Appendix).

This finding is consistent with the main finding about probability curvature in DetectGPT \citep{mitchell2023detectgpt}: human text is less likely to reach the local maxima of likelihood than model-generated text. In our case, the results suggest that there is more randomness in real human text, which makes it more stable against perturbation in time domain, such as the POS masking. 

\section{Conclusions}
In this study, we propose a new text detection method FourierGPT, which draws information from the spectrum view of relative likelihood scores in language, as the basis for distinguishing human and model texts. 
Our approach reaches better or competitive performances with state-of-the-art methods on typical zero-shot detection tasks, and particularly better on short text detection tasks. 

Our method has the following strengths: \underline{First}, it utilizes the \textbf{relative likelihood} ($z$-scores) rather than absolute values as used by most previous methods, which means it can capture likelihood patterns in language that are less dependent on the expressiveness of the generation model. We consider this as an advantage because the LM's capability of producing more ``likely'' texts inevitably grows, and thus, detection methods relying on the absolute ``thresholds'' of likelihood will also eventually fail.

\underline{Secondly}, we take a novel \textbf{spectrum view} of likelihood, which goes beyond the static view that simply aggregates likelihood at multiple time steps into a single value, but instead, characterizes the \emph{dynamic} features how likelihood changes in time. This spectrum view draws inspiration from cognitive characteristics of language production revealed in the psycholinguistics literature, such as UID, periodicity of surprisal etc. 
The likehood spectrum can reflect subtle differences in human and model languages that are otherwise indetectable.

\underline{Thirdly}, our method places a relatively low requirement on how accurate the likelihood scores need be estimated. A GPT-2 level model or even $n$-gram model suffices to provide likelihood features to reach a decent detection performance. 
It suggests that how the likelihood of human language distributes in time is a subtle process, which may not be easily mimicked by language models trained via maximum likelihood estimation. 
LLMs that try hard to squeeze out the gap between every single prediction and ground-truth token may still lack the ability to produce human-like language.


For future work, we will address the limitations with focus on: building stronger supervised classifiers by better utilizing the circularized spectrum; collecting larger datasets from broader domains and multiple languages; looking for more concrete linguistic cases (such as the ``\textit{Yes}''/``\textit{No}'' example in \Cref{sec:yesno}) to provide richer interpretations for the spectrum-view of likelihood.

\section{Limitations}
The limitations of the current study are: \underline{First}, the pair-wise classifier requires that the two texts being classified must be generated from the same prompt. It is yet to be verified whether the classifier's performance will retain if the source prompts are different. 
\underline{Second}, the supervised classifier is not a strict zero-shot detector, and its performance still has space for improvement. We argue that it would be less of an issue if the classifier learns the general features instead of idiosyncrasy in certain data. This, however, requires further investigation to whether likelihood spectrum is such a general feature. 
\underline{Third}, the datasets examined are relatively small. It is worth exploration on larger datasets (especially short text corpus, such as QA) to further confirm the effectiveness of the method. 

\paragraph{Acknowledgments}
We sincerely thank all the reviewers for their efforts in pointing out the weakness in the paper and their insightful advice for future improvement.
Yu Wang is supported by the \href{https://www.dfg.de}{Deutsche For\-schungs\-gemeinschaft (DFG)}: \href{https://gepris.dfg.de/gepris/projekt/438445824}{TRR 318/1 2021 – 438445824}.



\bibliography{custom}

\newpage
\appendix

\section{Hyperparameters for Supervised Classifiers}
\label{sec:appendix classification}

Detail classification results are shown on \Cref{tab:bigram-clf} and \Cref{tab:spec-clf}. The result is the mean of 5-fold cross-validation score of each dataset.

\begin{table*}[h]
\centering
    \begin{tabular}{l|l|c|c|c|c|c|c}
\toprule
Dataset & Gen. model & HGBT & KNN & MLP & SVM & NB & LR \\
\hline
\multirow{3}*{PubMed}   & GPT4  & 0.567 & 0.580 & 0.580 & 0.593 & 0.573 & 0.597 \\
                        & GPT3.5& 0.597 & 0.583 & 0.597 & 0.607 & 0.607 & 0.603 \\
                        & GPT3  & 0.577 & 0.600 & 0.613 & 0.603 & 0.597 & 0.593 \\
\hline
\multirow{3}*{Writing}  & GPT4  & 0.663 & 0.677 & 0.707 & 0.717 & 0.710 & 0.677 \\
                        & GPT3.5& 0.680 & 0.713 & 0.693 & 0.743 & 0.733 & 0.737 \\
                        & GPT3  & 0.553 & 0.543 & 0.537 & 0.530 & 0.530 & 0.527 \\
\hline
\multirow{3}*{Xsum}     & GPT4  & 0.693 & 0.670 & 0.713 & 0.717 & 0.707 & 0.697 \\
                        & GPT3.5& 0.640 & 0.623 & 0.660 & 0.677 & 0.660 & 0.667 \\
                        & GPT3  & 0.557 & 0.560 & 0.550 & 0.557 & 0.550 & 0.563 \\
\bottomrule
    \end{tabular}
    \label{tab:bigram-clf}
    \caption{Accuracy of supervised classifier using likelihood spectrum estimated by bigram language model.}
\end{table*}

\begin{table*}[h]
\centering
    \begin{tabular}{l|l|c|c|c|c|c|c}
\toprule
Dataset & Gen. model & HGBT & KNN & MLP & SVM & NB & LR \\
\hline
\multirow{3}*{PubMed}   & GPT4  & 0.797 & 0.800 & 0.800 & 0.827 & 0.806 & 0.810 \\
                        & GPT3.5& 0.580 & 0.583 & 0.557 & 0.600 & 0.533 & 0.573 \\
                        & GPT3  & 0.580 & 0.553 & 0.553 & 0.570 & 0.567 & 0.557 \\
\hline
\multirow{3}*{Writing}  & GPT4  & 0.713 & 0.690 & 0.693 & 0.707 & 0.717 & 0.663 \\
                        & GPT3.5& 0.687 & 0.683 & 0.713 & 0.737 & 0.750 & 0.723 \\
                        & GPT3  & 0.797 & 0.800 & 0.817 & 0.827 & 0.807 & 0.810 \\
\hline
\multirow{3}*{Xsum}     & GPT4  & 0.717 & 0.690 & 0.737 & 0.740 & 0.733 & 0.723 \\
                        & GPT3.5& 0.730 & 0.710 & 0.727 & 0.750 & 0.737 & 0.730 \\
                        & GPT3  & 0.717 & 0.687 & 0.723 & 0.740 & 0.733 & 0.723 \\
\bottomrule
    \end{tabular}
    \label{tab:spec-clf}
    \caption{Accuracy of supervised classifier using likelihood spectrum estimated by GPT2-xl.}
\end{table*}

For classification, the data will pass through a scaler, and then a k-best feature selector, at last, the classifier. We apply grid-search on different parameters and report the best outputs. The parameters of the overall workflow are shown below:
\begin{itemize}\setlength\itemsep{0em}
    \item Scaler: MinMax, ZScore, Robust
    \item KBestFeatures: 50, 80, 100, 120, 150, 200, 250, 300, 400, 500
    \item SVM (Support Vector Machine): 
    \begin{itemize}
        \item kernel: rbf, linear
        \item C: 1, 2, 10
        \item gamma: scale, auto
    \end{itemize}
    \item HGBT (Histogram Gradient Boosting Trees):
    \begin{itemize}
        \item max iter: 500
        \item learning rate: 0.1, 0.05, 0.01, 0.005, 0.001
        \item min samples leaf: 7, 13
    \end{itemize}
    \item MLP (Multi-Layer Perceptrons):
    \begin{itemize}
        \item constant learning rate: 0.001
        \item SGD momentum: 0.9
        \item max iter: 800
        \item hidden layer: (500), (500, 50)
    \end{itemize}
    \item LR (Logistic Regression):
    \begin{itemize}
        \item solver: liblinear
        \item penalty: l1, l2
        \item C: 1, 2, 10
    \end{itemize}
    \item KNN (K-Neighbors Classifier):
    \begin{itemize}
        \item n: 3, 5, 7, 9
    \end{itemize}
    \item NB (Complement Naive Bayes):
    \begin{itemize}
        \item alpha: 0.5, 1, 2
    \end{itemize}
\end{itemize}

\begin{table*}[h]
\small
\centering
\begin{tabular}{lllllllllll} 
\hline
~    &   & Pubmed &         &        & Writing &         &        & Xsum   &         &         \\ 
\cline{3-11}
     &   & GPT-3  & GPT-3.5 & GPT-4  & GPT-3   & GPT-3.5 & GPT-4  & GPT-3  & GPT-3.5 & GPT-4   \\ 
\hline
VERB & H & 0.8559 & 0.8574  & 0.8536 & 0.8155  & 0.8185  & 0.8187 & 0.8062 & 0.8015  & \textbf{0.8017}  \\
     & M & 0.8497 & 0.8390  & 0.8311 & 0.8001  & 0.7950  & 0.8074 & 0.8036 & 0.7947  & \textbf{0.8029}  \\ 
\hline
NOUN & H & 0.7564 & 0.7598  & 0.7564 & 0.7972  & 0.7982  & 0.7979 & 0.7736 & 0.7717  & 0.7714  \\
     & M & 0.7528 & 0.7280  & 0.7267 & 0.7827  & 0.7624  & 0.7676 & 0.7639 & 0.7537  & 0.7600  \\ 
\hline
ADJ  & H & \textbf{0.8109} & 0.8110  & 0.8077 & 0.8700  & 0.8679  & 0.8674 & 0.8575 & 0.8526  & 0.8548  \\
     & M & \textbf{0.8196} & 0.7830  & 0.7833 & 0.8485  & 0.8349  & 0.8357 & 0.8486 & 0.8272  & 0.8302  \\ 
\hline
NVA  & H & 0.6976 & 0.6961  & \textbf{0.6953} & 0.7397  & 0.7433  & 0.7414 & 0.7250 & 0.7238  & 0.7232  \\
     & M & 0.6938 & 0.6634  & \textbf{0.6697} & 0.7249  & 0.7151  & 0.7207 & 0.7220 & 0.7083  & 0.7165  \\
\hline
\end{tabular}
\caption{Attention mask effect on Spectral Overlap. H denotes human. M denotes model.Bolded Number is the minority which shows that likelihood spectrum from human text changes more than the equivalent from model generated text after attention mask. All the other number shows the opposite situation.}
\label{tab:speclap}
\end{table*}

\begin{figure}[h]
	
	\begin{minipage}{0.32\linewidth}
		\vspace{3pt}
		\centerline{\includegraphics[width=\textwidth]{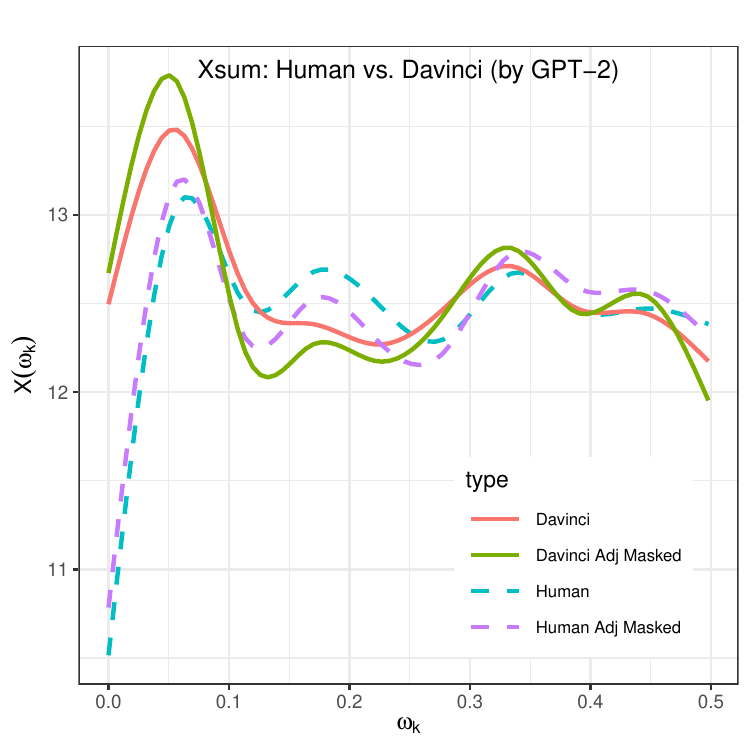}}
		\centerline{Xsum}
	\end{minipage}
	\begin{minipage}{0.32\linewidth}
		\vspace{3pt}
		\centerline{\includegraphics[width=\textwidth]{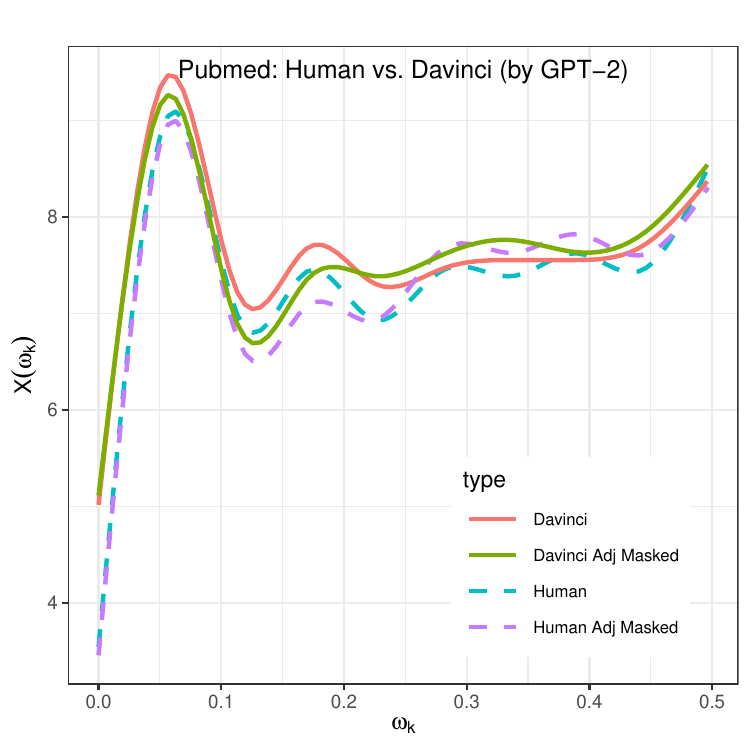}} 
		\centerline{PubMed}
	\end{minipage}
	\begin{minipage}{0.32\linewidth}
		\vspace{3pt}
		\centerline{\includegraphics[width=\textwidth]{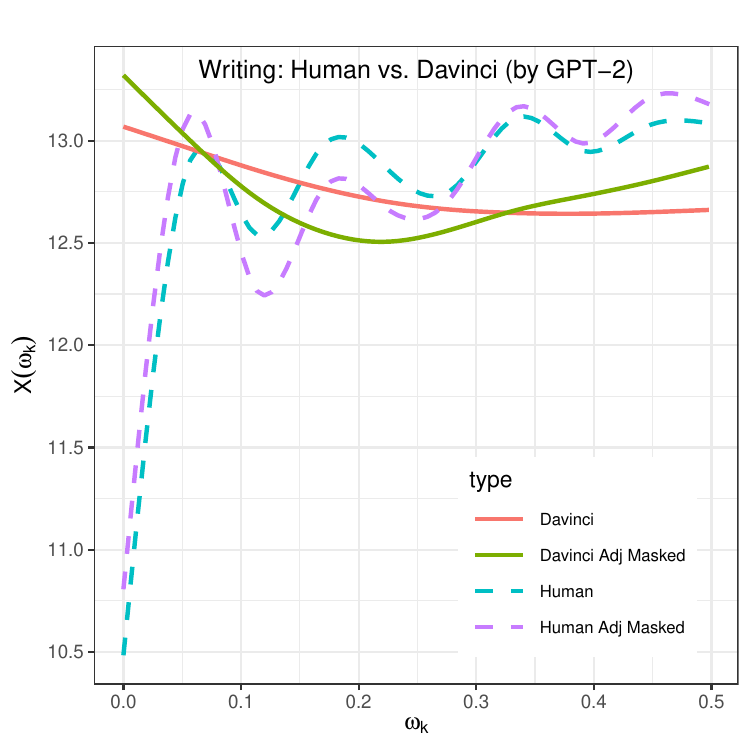}}
		\centerline{Writing}
	\end{minipage}
	\caption{Likelihood spectrum before and after attention mask on ADJ with \textbf{GPT-2} estimator.}
	\label{fig:MASK-adj-gpt2}
\end{figure}

\begin{figure}[h]
	
	\begin{minipage}{0.32\linewidth}
		\vspace{3pt}
		\centerline{\includegraphics[width=\textwidth]{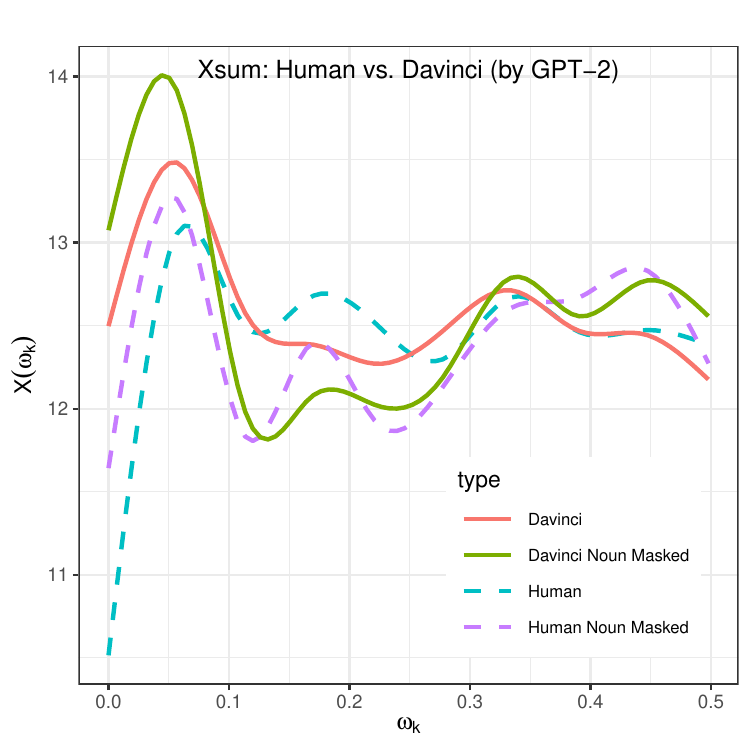}}
		\centerline{Xsum}
	\end{minipage}
	\begin{minipage}{0.32\linewidth}
		\vspace{3pt}
		\centerline{\includegraphics[width=\textwidth]{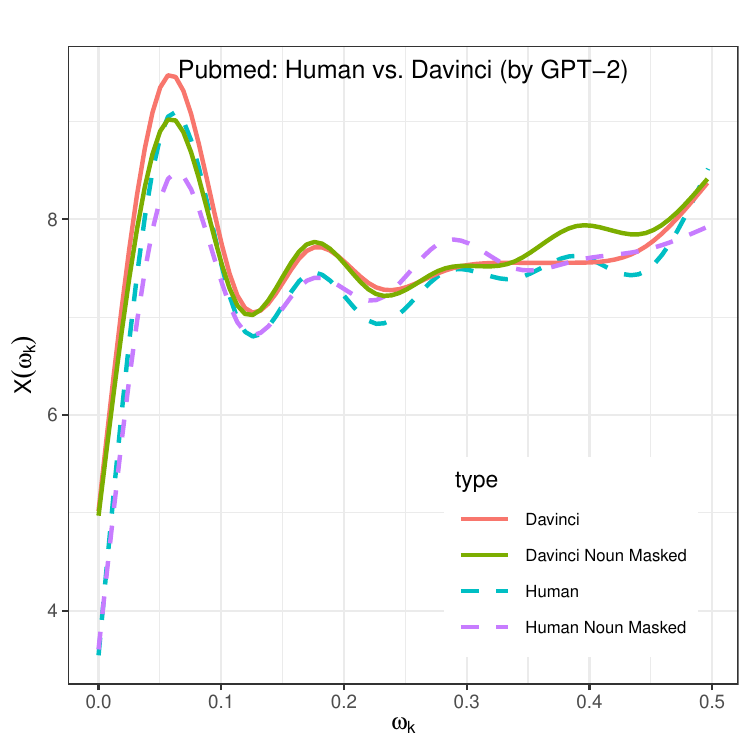}} 
		\centerline{PubMed}
	\end{minipage}
	\begin{minipage}{0.32\linewidth}
		\vspace{3pt}
		\centerline{\includegraphics[width=\textwidth]{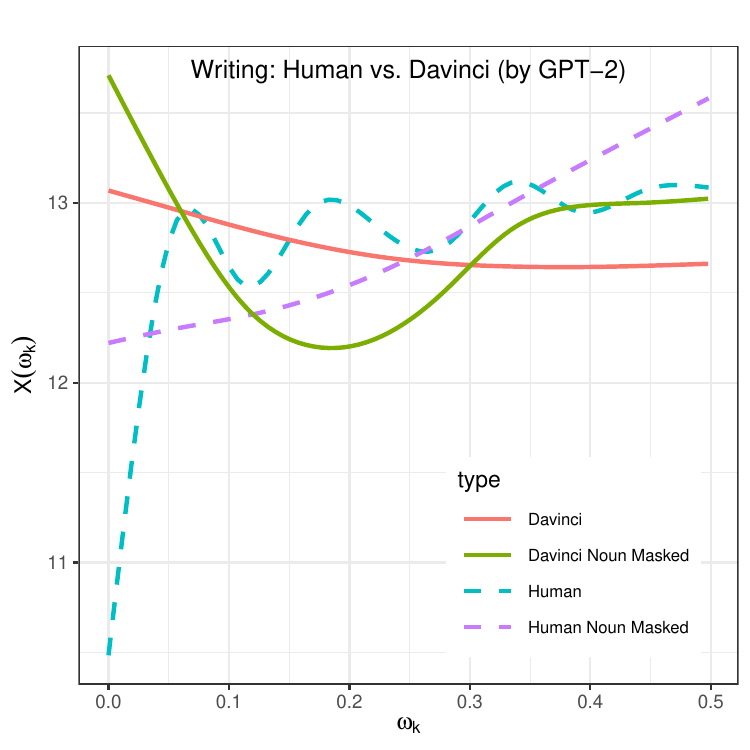}}
		\centerline{Writing}
	\end{minipage}
	\caption{Likelihood spectrum before and after attention mask on NOUN with \textbf{GPT-2} estimator.}
	\label{fig:MASK-noun-gpt2}
\end{figure}

\begin{figure}[h]
	
	\begin{minipage}{0.32\linewidth}
		\vspace{3pt}
		\centerline{\includegraphics[width=\textwidth]{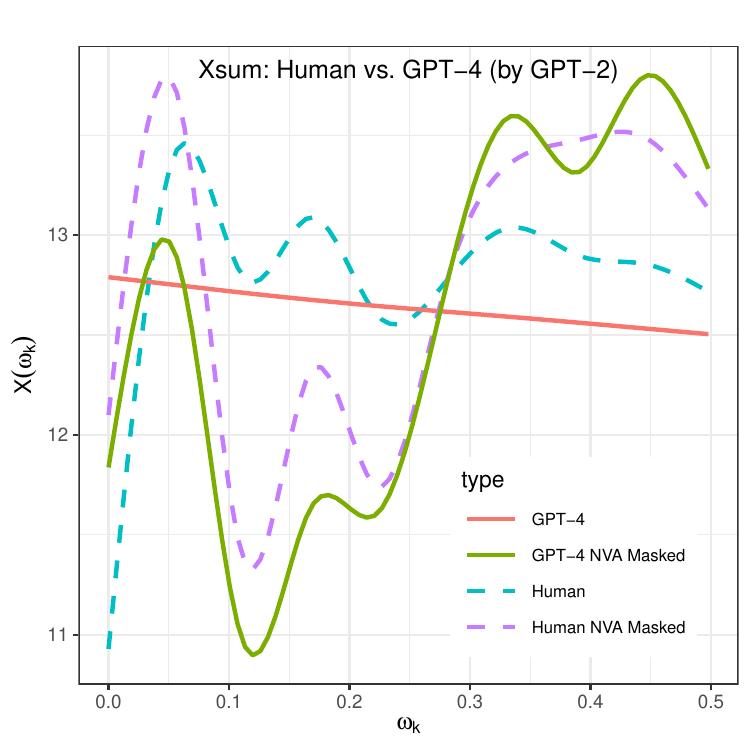}}
		\centerline{Xsum}
	\end{minipage}
	\begin{minipage}{0.32\linewidth}
		\vspace{3pt}
		\centerline{\includegraphics[width=\textwidth]{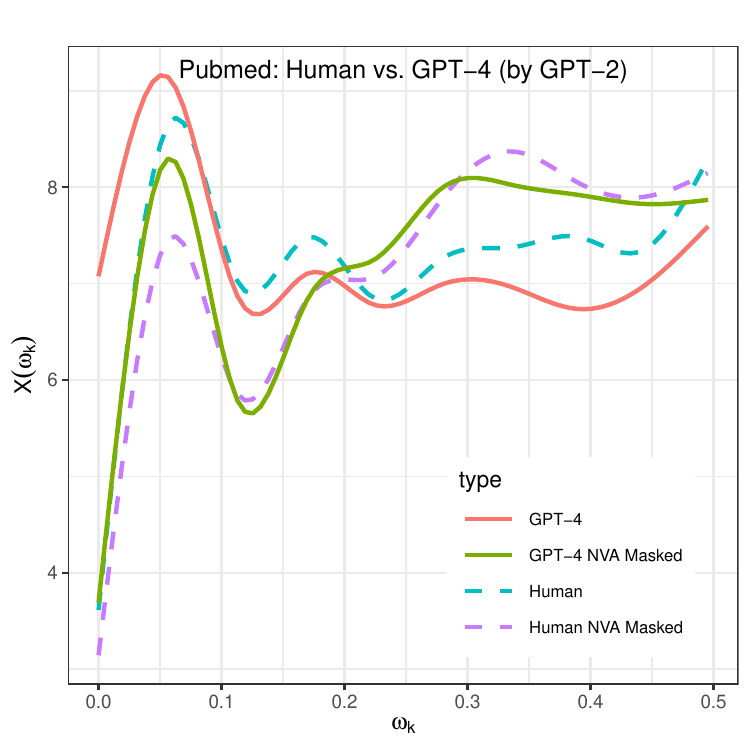}} 
		\centerline{PubMed}
	\end{minipage}
	\begin{minipage}{0.32\linewidth}
		\vspace{3pt}
		\centerline{\includegraphics[width=\textwidth]{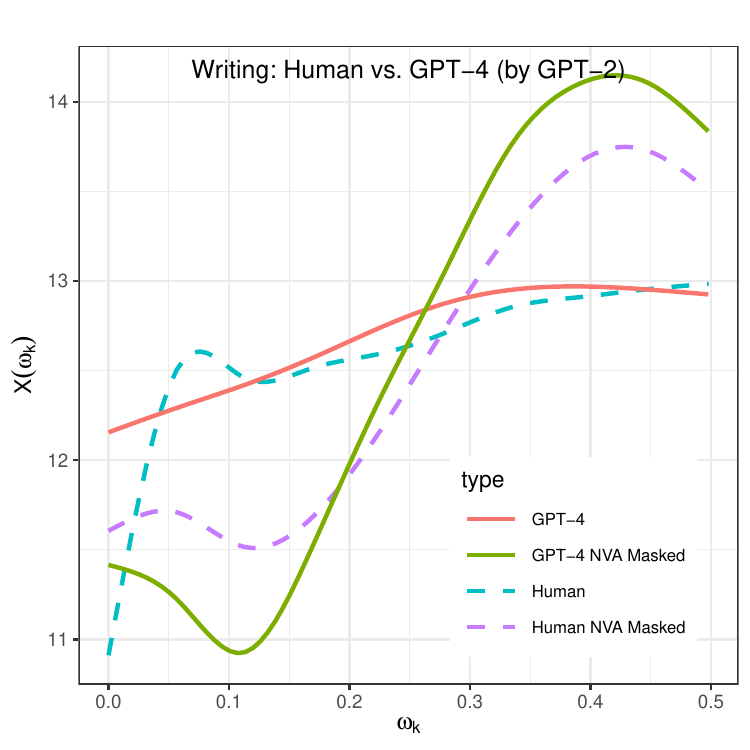}}
		\centerline{Writing}
	\end{minipage}
	\caption{Likelihood spectrum before and after attention mask on NVA with \textbf{GPT-2} estimator.}
	\label{fig:MASK-NVA-gpt2}
\end{figure}

\begin{figure}[h]
	
	\begin{minipage}{0.32\linewidth}
		\vspace{3pt}
		\centerline{\includegraphics[width=\textwidth]{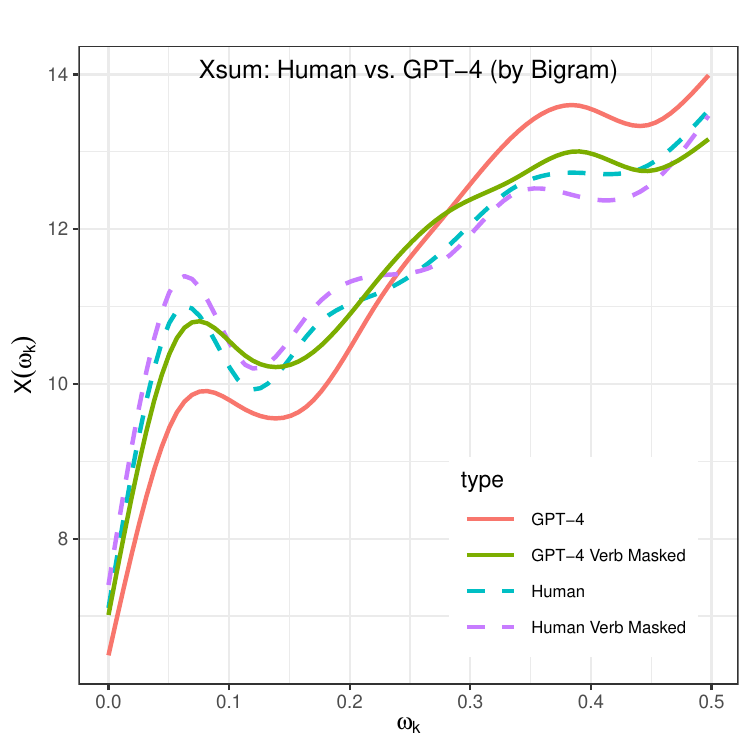}}
		\centerline{Xsum}
	\end{minipage}
	\begin{minipage}{0.32\linewidth}
		\vspace{3pt}
		\centerline{\includegraphics[width=\textwidth]{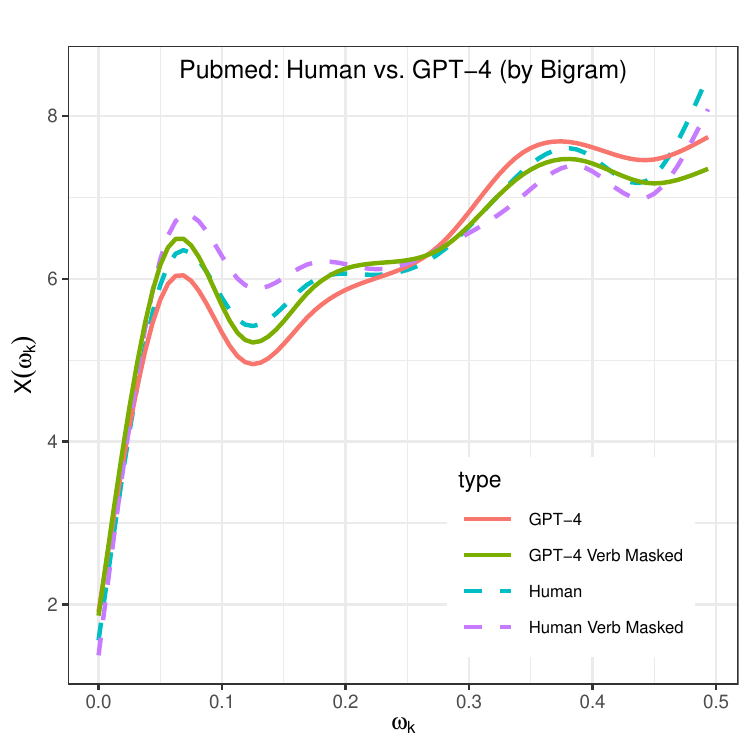}} 
		\centerline{PubMed}
	\end{minipage}
	\begin{minipage}{0.32\linewidth}
		\vspace{3pt}
		\centerline{\includegraphics[width=\textwidth]{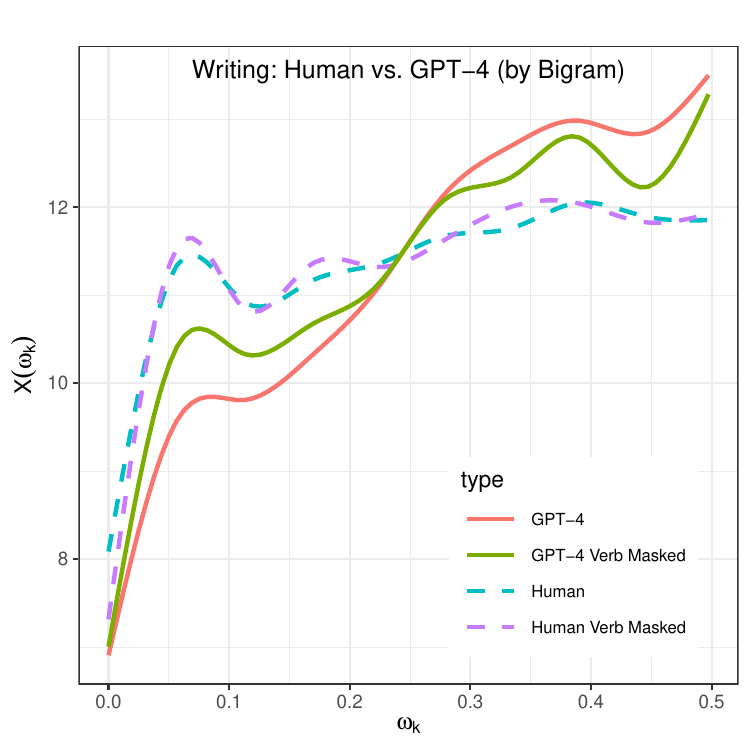}}
		\centerline{Writing}
	\end{minipage}
 
	\caption{Likelihood spectrum before and after attention mask on VERB with \textbf{bigram} estimator.}
	\label{fig:MASK-verb-bi}
\end{figure}

\begin{figure}[h]
	
	\begin{minipage}{0.32\linewidth}
		\vspace{3pt}
		\centerline{\includegraphics[width=\textwidth]{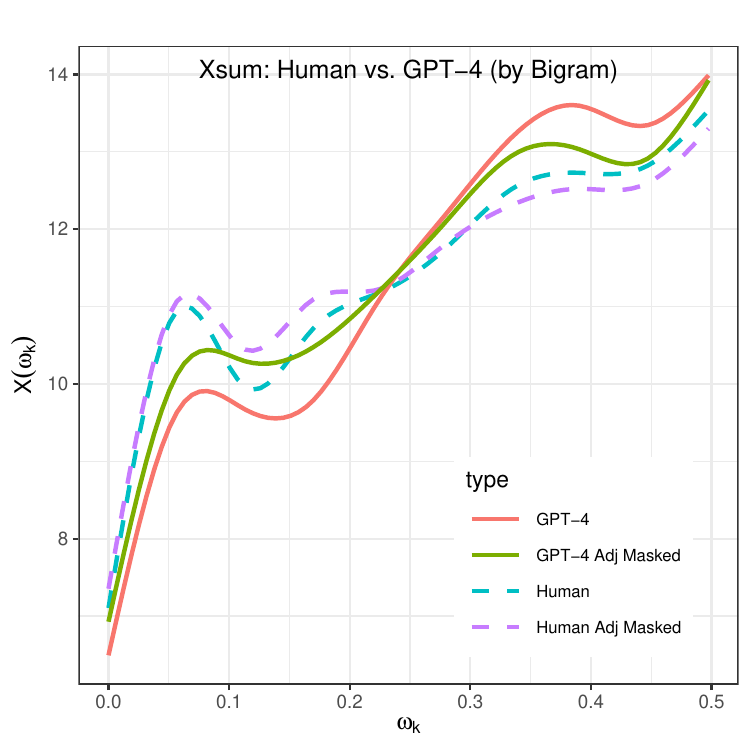}}
		\centerline{Xsum}
	\end{minipage}
	\begin{minipage}{0.32\linewidth}
		\vspace{3pt}
		\centerline{\includegraphics[width=\textwidth]{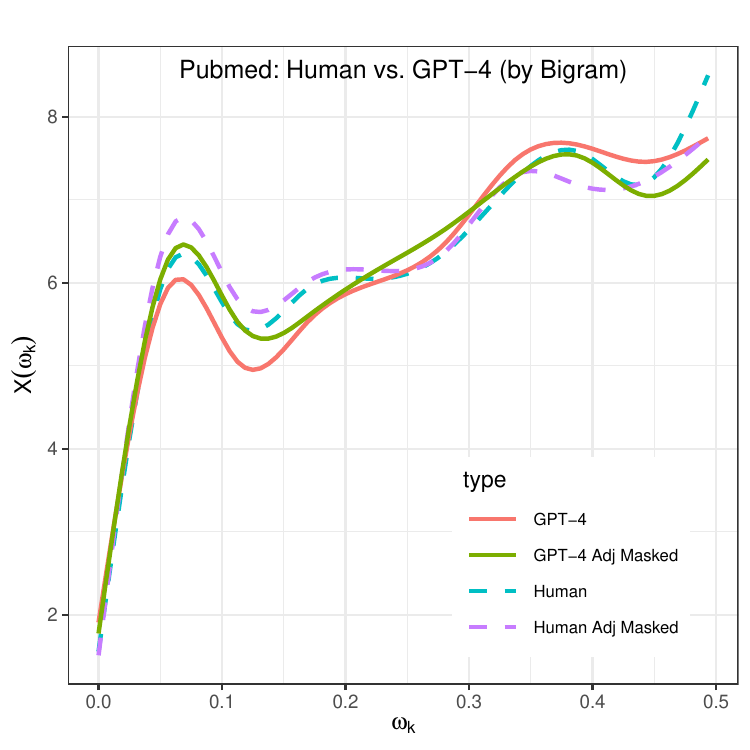}} 
		\centerline{PubMed}
	\end{minipage}
	\begin{minipage}{0.32\linewidth}
		\vspace{3pt}
		\centerline{\includegraphics[width=\textwidth]{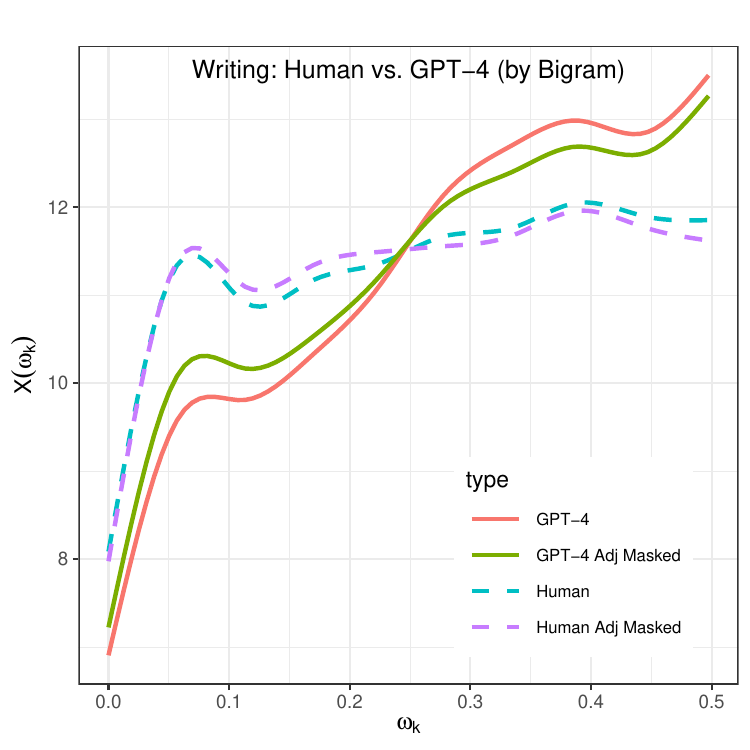}}
		\centerline{Writing}
	\end{minipage}
 
	\caption{Likelihood spectrum before and after attention mask on ADJ with \textbf{bigram} estimator.}
	\label{fig:MASK-adj-bi}
\end{figure}

\begin{figure}[h]
	
	\begin{minipage}{0.32\linewidth}
		\vspace{3pt}
		\centerline{\includegraphics[width=\textwidth]{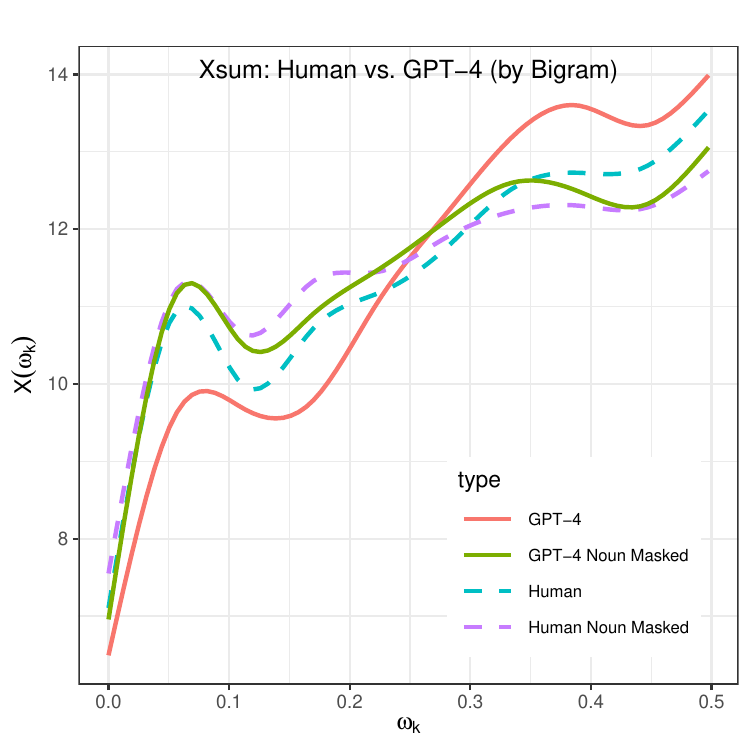}}
		\centerline{Xsum}
	\end{minipage}
	\begin{minipage}{0.32\linewidth}
		\vspace{3pt}
		\centerline{\includegraphics[width=\textwidth]{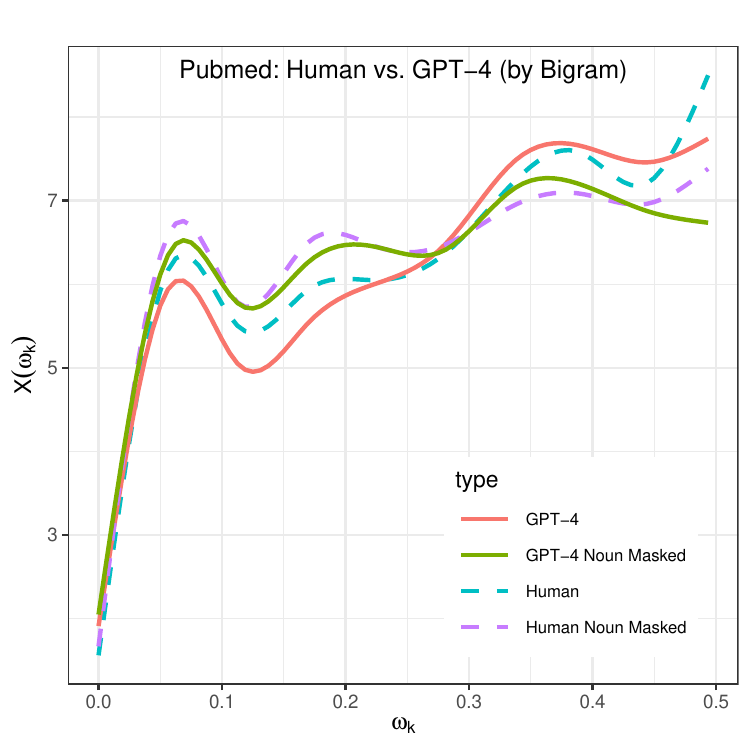}} 
		\centerline{PubMed}
	\end{minipage}
	\begin{minipage}{0.32\linewidth}
		\vspace{3pt}
		\centerline{\includegraphics[width=\textwidth]{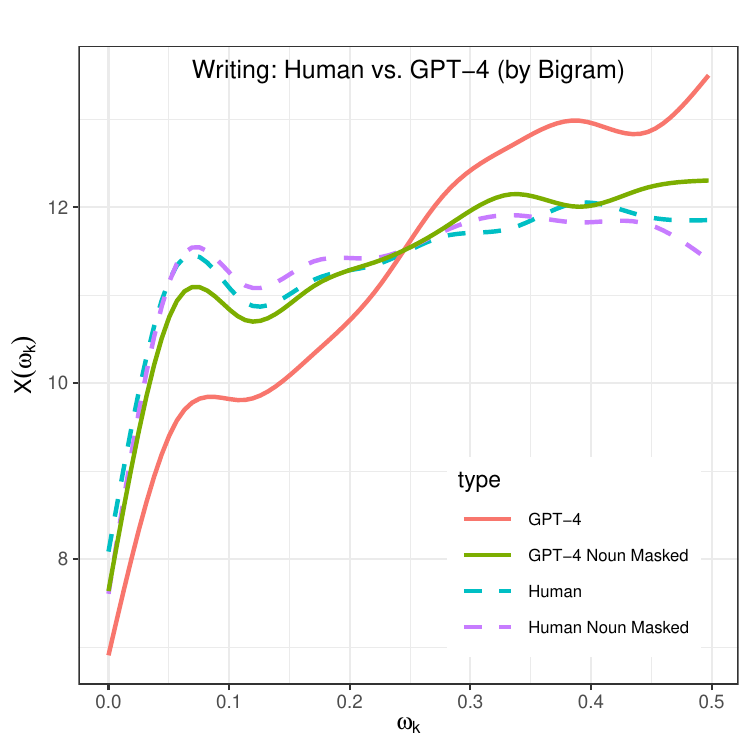}}
		\centerline{Writing}
	\end{minipage}
 
	\caption{Likelihood spectrum before and after attention mask on NOUN with \textbf{bigram} estimator.}
	\label{fig:MASK-noun-bi}
\end{figure}

\begin{figure}[h]
	\begin{minipage}{0.32\linewidth}
		\vspace{3pt}
		\centerline{\includegraphics[width=\textwidth]{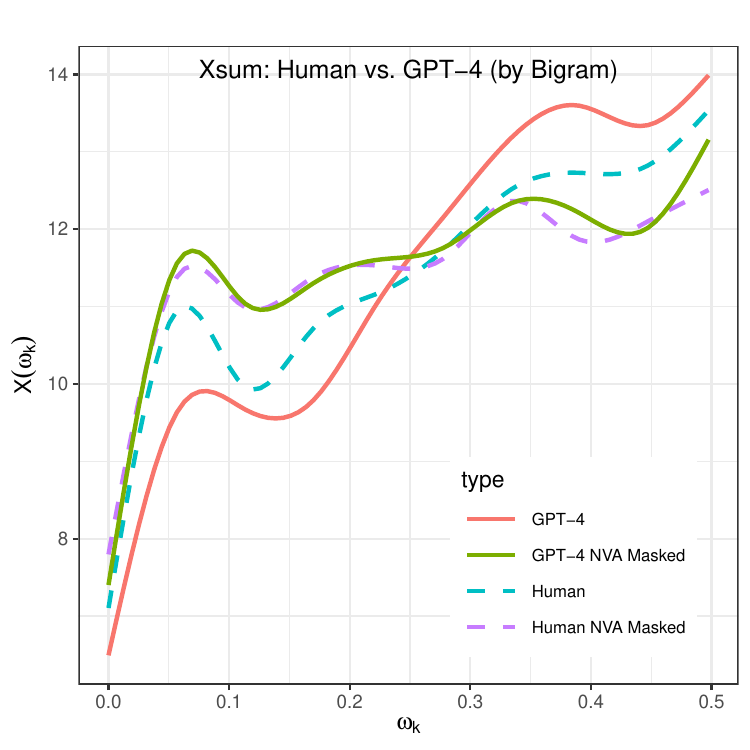}}
		\centerline{Xsum}
	\end{minipage}
	\begin{minipage}{0.32\linewidth}
		\vspace{3pt}
		\centerline{\includegraphics[width=\textwidth]{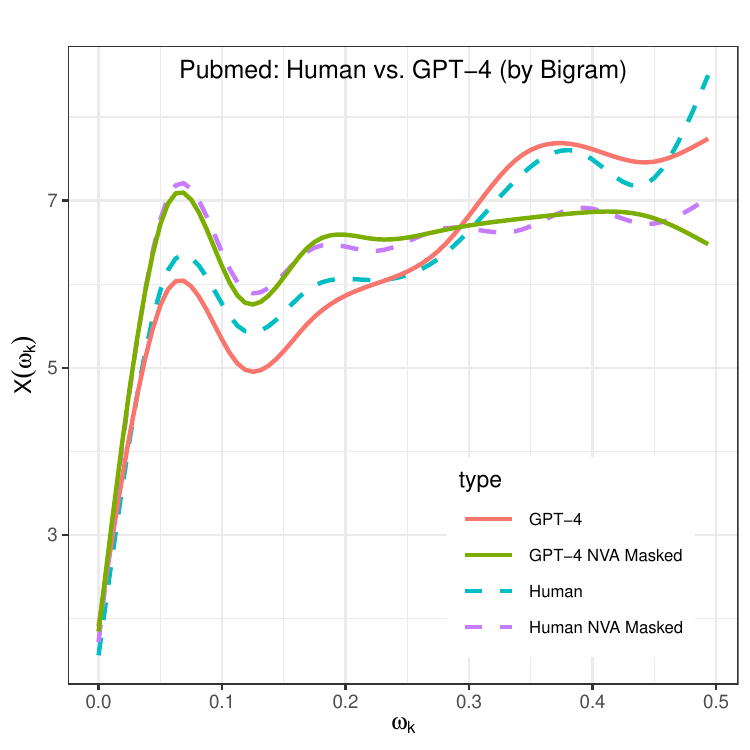}} 
		\centerline{PubMed}
	\end{minipage}
	\begin{minipage}{0.32\linewidth}
		\vspace{3pt}
		\centerline{\includegraphics[width=\textwidth]{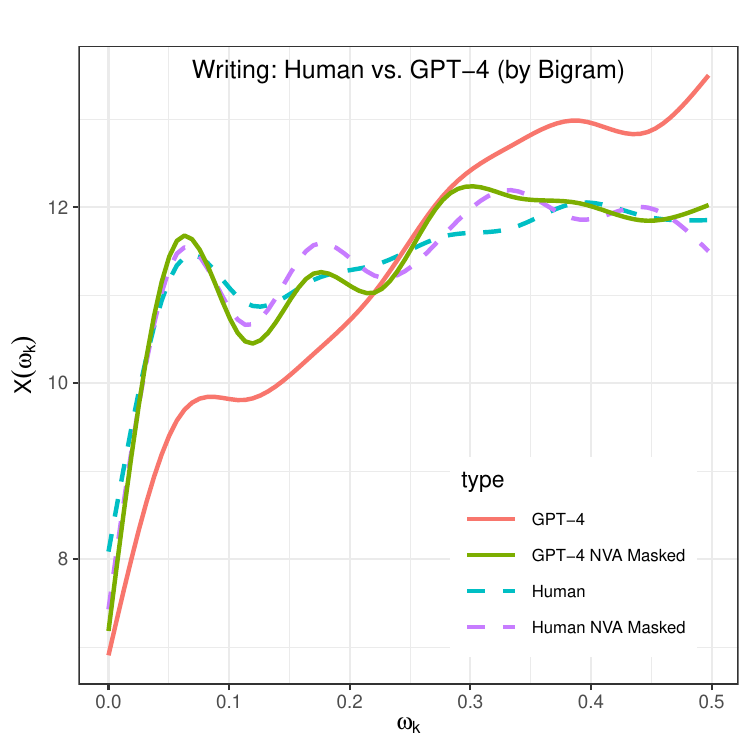}}
		\centerline{Writing}
	\end{minipage}
 
	\caption{Likelihood spectrum before and after attention mask on NVA with \textbf{bigram} estimator.}
	\label{fig:MASK-NVA-bi}
\end{figure}

\end{document}